\documentclass[runningheads]{llncs}
\usepackage[T1]{fontenc}
\usepackage{graphicx}
\usepackage{amsmath}
\usepackage{amssymb}
\usepackage{array}
\usepackage{cellspace}
\usepackage{adjustbox}
\usepackage{comment}
\graphicspath{ {./images/} }
\usepackage{booktabs}
\usepackage{bm}
\usepackage{textcomp}
\begin{document}
%
% \title{A Survey on Image Vectorization\thanks{Supported by organization x.}}
\title{Image Vectorization: a Review}
% \author{Anonymous Submission}
\author{Maria Dziuba\inst{1}\and Ivan Jarsky\inst{1}\and Valeria Efimova\inst{1}\and Andrey Filchenkov\inst{2}}
\authorrunning{M. Dziuba et al.}
\institute{
ITMO University \email{\{dziuba.maria,ivanjarsky,vefimova\}@itmo.ru}
\and
GO AI LAB \email{aaafil@gmail.com}
}
\maketitle              % typeset the header of the contribution
\begin{abstract}
% The abstract should briefly summarize the contents of the paper in 150--250 words.

Nowadays, there are many diffusion and autoregressive models that show impressive results for generating images from text and other input domains. 
However, these methods are not intended for ultra-high-resolution image synthesis. 
Vector graphics are devoid of this disadvantage, so the generation of images in this format looks very promising. 
Instead of generating vector images directly, you can first synthesize a raster image and then apply vectorization. 
Vectorization is the process of converting a raster image into a similar vector image using primitive shapes. 
Besides being similar, generated vector image is also required to contain the minimum number of shapes for rendering. 
In this paper, we focus specifically on machine learning-compatible vectorization methods. We are considering Mang2Vec, Deep Vectorization of Technical Drawings, DiffVG, and LIVE models. We also provide a brief overview of existing online methods. We also recall other algorithmic methods, Im2Vec and ClipGEN models, but they do not participate in the comparison, since there is no open implementation of these methods or their official implementations do not work correctly.
Our research shows that despite the ability to directly specify the number and type of shapes, existing machine learning methods work for a very long time and do not accurately recreate the original image. We believe that there is no fast universal automatic approach and human control is required for every method.

% \keywords{First keyword  \and Second keyword \and Another keyword.}
\end{abstract}
\section{Introduction}

In computer graphics, two main approaches for image representation coexist. 
While a bitmap image is a matrix of pixels, a vector image is a sequence of shapes drawn with the canvas. 
Raster graphics are commonly used for complex images containing a large number of visual details and complex color transitions. 
Most often these are photos and photorealistic drawings. 
At the same time, vector images consist of figures, which typically have a constant one-color fill. 
This results in the simplicity and abstractness of the resulting image. 
Therefore, the primary domains of vector graphics are icons, logos, simple illustrations, and fonts. 
Vector images are easily embedded in HTML markup and a crucial requirement is the small size of the code describing them for fast data transfer to the user and subsequent rapid rendering. 
The most popular vector format is SVG, which defines a vector image as a tag sequence using XML markup. 
Each tag can use XML attributes to specify the shape color characteristics and transformations. 
Using this markup, a renderer program draws an image consisting of the specified figures. 
Thus, the task of vectorizing a bitmap image is similar to the task of obtaining a sequence of shapes and their parameters that together form the original raster image.

In 2014, generative adversarial models~\cite{goodfellow2020generative} became the first machine learning algorithm for image synthesis. 
Since then, image synthesis has become an important part of digital art, data augmentation techniques, design, fashion, and several other domains.
Currently, the image generation task is solved with deep generative models based on diffusion models~\cite{rombach2022high,saharia2022photorealistic} and autoregressive models~\cite{esser2021taming,yu2022scaling}.

%AF
Modern generative methods have recently achieved significant success in the raster domain. 
However, despite these approaches being designed to generate highly realistic images in different styles from the input text, the resulting images do not have a high resolution; usually, it is less than 2048x2048 pixels. 
However, it is not enough for logos, covers, and for printing.
In these domains, vector graphics are standard.

One of the ways to obtain such images is to use vector graphics instead of raster graphics. 
Notably, little research is done in vector image generation~\cite{deep-svg,wang2021deepvecfont,clipdraw,efimova2022conditional,svg-vae,schaldenbrand2022styleclipdraw,jain2022vectorfusion,efimova2023neural}.
The creation of vector images is still done by humans, but working with the vector image code is not an easy task. 
Therefore, the ability to generate such images automatically with a minimal number of post-processing steps is very necessary. 

A possible solution for obtaining a vector image is the vectorization of raster images. 
%AF
It may also allow the enrichment of vector datasets that are required for training vector image generation algorithms, but the size of which is still not sufficient in comparison with bitmap image datasets.

The existing image vectorization methods can be divided into two categories: algorithmic and machine learning-based methods. 
Algorithmic approaches have recently been reviewed~\cite{tian2022survey}. The authors classify the vectorization methods as mesh-based and curve-based. 
Mesh-based methods split the image into non-overlapping 2D patches and interpolate colors across them. The patch shape can be triangular~\cite{zhou2014representing,liao2012subdivision,hettinga2021efficient}, rectangular~\cite{price2006object,sun2007image,lai2009automatic}, or even irregular, for example, in the form of B\'ezigons, closed regions bounded by B\'ezier curves~\cite{Swaminarayan2006RapidAP,Lecot2006ArdecoAR,Yang2016EffectiveCI}, and the patch vertices or interior can store color and other attributes. 
Curve-based methods are based on diffusion curves, which are B\'ezier curves with colors defined on their left and right sides.
Color discontinuities can be modeled by diffusing the colors from both sides of the curves to create the resulting image. 
For smooth edges, the diffusion process might be followed by a blurring phase. There are different formulations of diffusion curves to work with: basic and harmonic~\cite{Dai2013AutomaticIV,zhao2017inverse,jeschke2011estimating}, and biharmonic~\cite{xie2014hierarchical} as well.

% The majority of recent work on extracting curve-based representations from pictures has been directed toward diffusion curves, particularly the original, harmonic, and biharmonic formulations. 
% Because the original formulation and the harmonic version extract the same attributes, the authors discuss them together. 
% Unfortunately in this paper there's no comparison between generated images or other attempts to evaluate the results.

% The main conclusion about vectorization is that ... . % Он вообще размытый какой-то 

% Tian et al. have made a complete and comprehensive review of image vectorization by algorithmic methods. 
% They divided the methods into mesh-based and curve based. 

Although several vectorization methods exist, no decent comparison was made to the best of our knowledge. 
In this paper, we focus on machine learning-compatible image vectorization methods. 
We aim to classify and compare them using different evaluation criteria. 
The main comparison criteria of the vectorizing methods are:
1) similarity to the original bitmap; 
2) the simplicity or complexity of the resulting image including the number of shapes and their parameters; 
3) the speed of generation; 
4) versatility~--- the ability to generate a fairly accurate copy of the input image without prior model training; 
5) human control to adjust hyperparameters.

The contributions of this paper are overview of machine learning-compatible vectorization methods and comparison of their performance.

\section{Machine Learning-compatible Methods}

\subsection{DiffVG}

\begin{figure}[h]
\begin{center}
    \begin{tabular}{
    >{\centering\arraybackslash}m{1.5cm} |
    >{\centering\arraybackslash}m{1.5cm}
    >{\centering\arraybackslash}m{1.5cm}
    >{\centering\arraybackslash}m{1.5cm}
    >{\centering\arraybackslash}m{1.5cm}
    >{\centering\arraybackslash}m{1.5cm}
    >{\centering\arraybackslash}m{1.5cm}
    }
    \includegraphics[width=1.5cm, height=1.5cm]{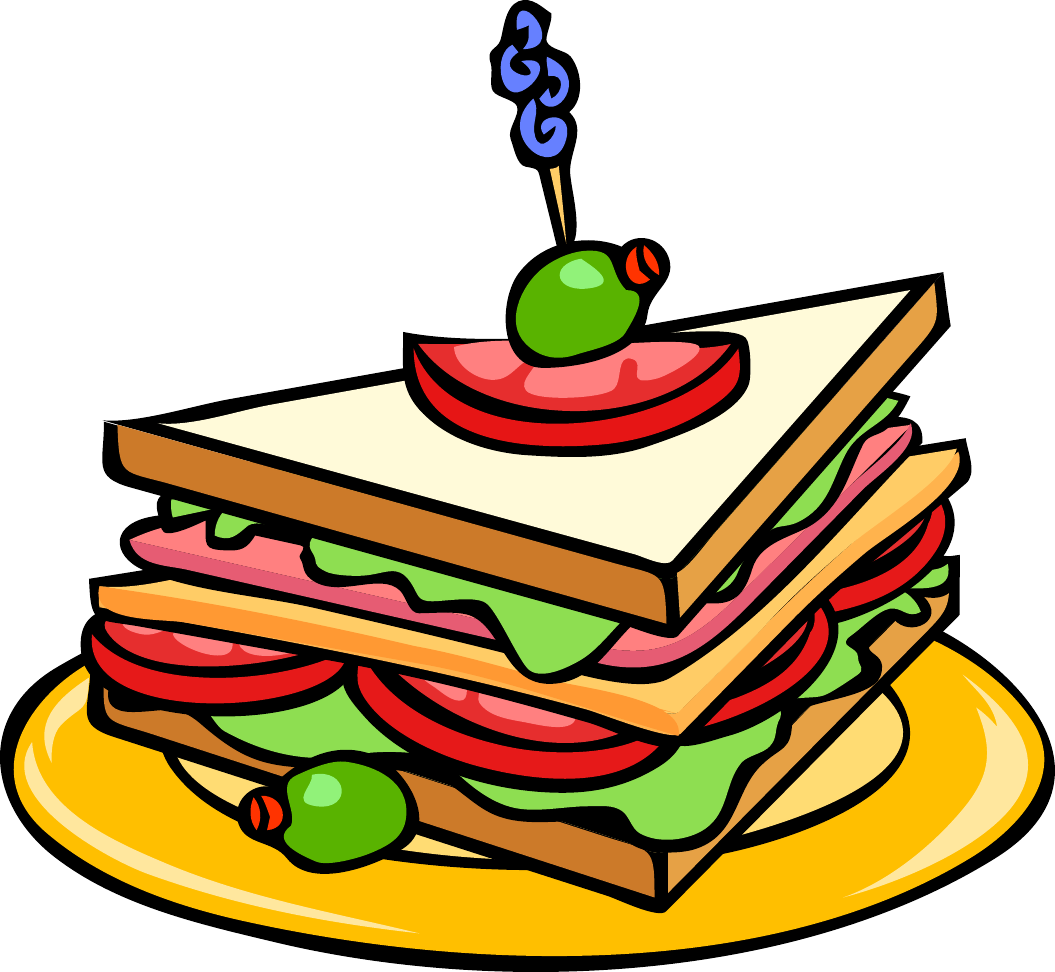} &
    \includegraphics[width=1.5cm, height=1.5cm]{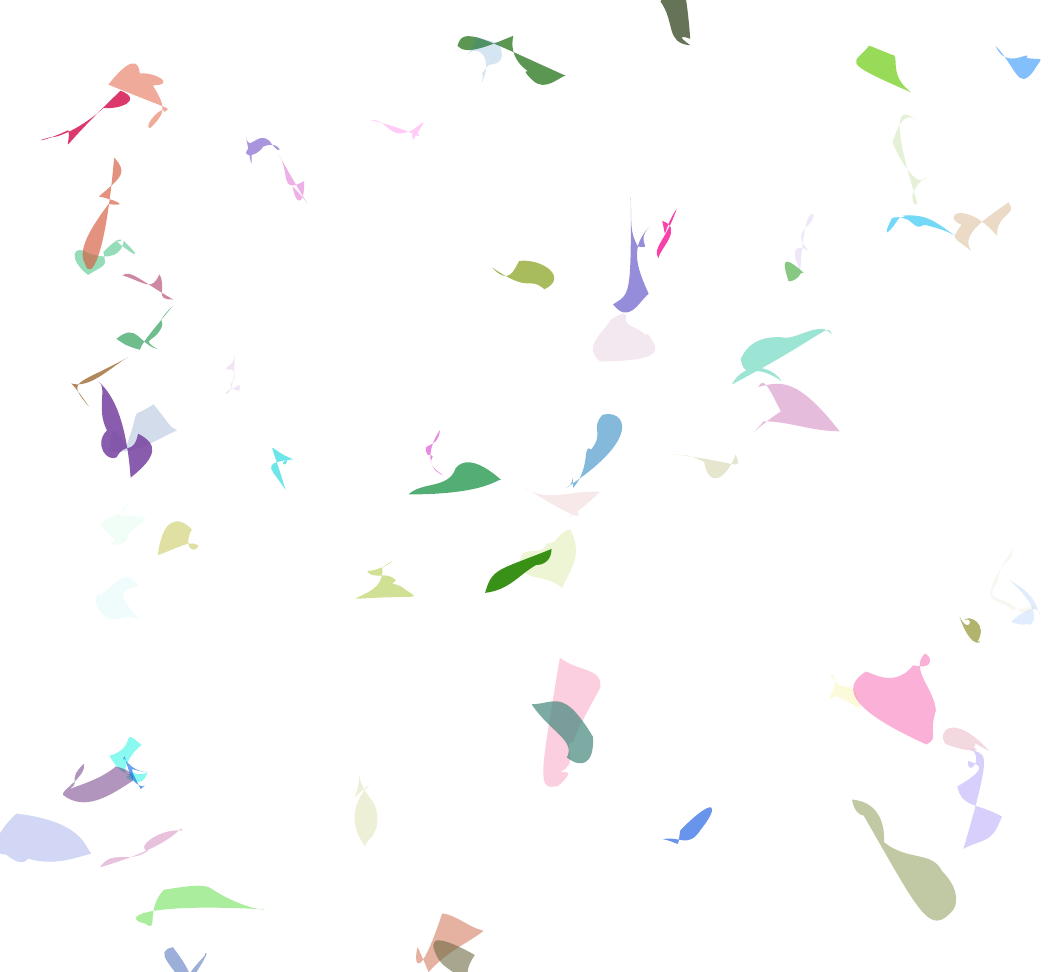} &
    \includegraphics[width=1.5cm, height=1.5cm]{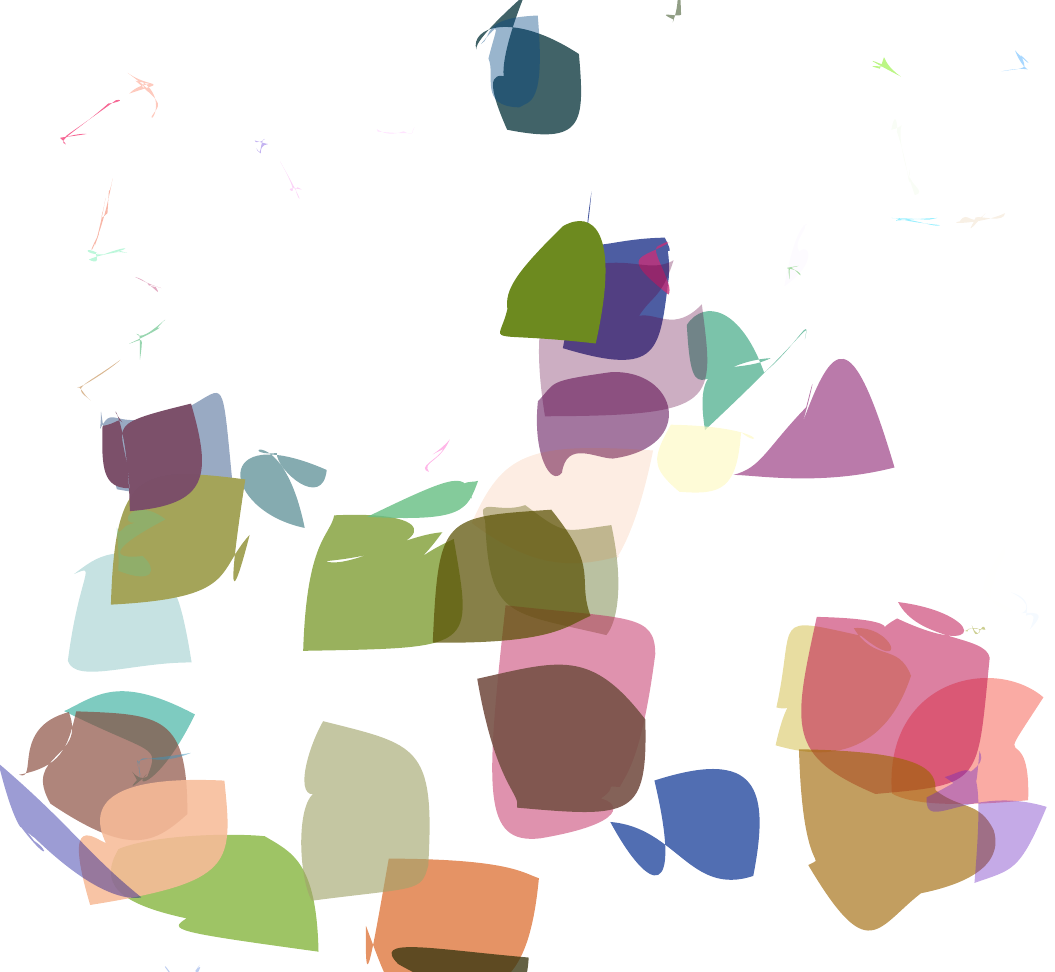} &
    \includegraphics[width=1.5cm, height=1.5cm]{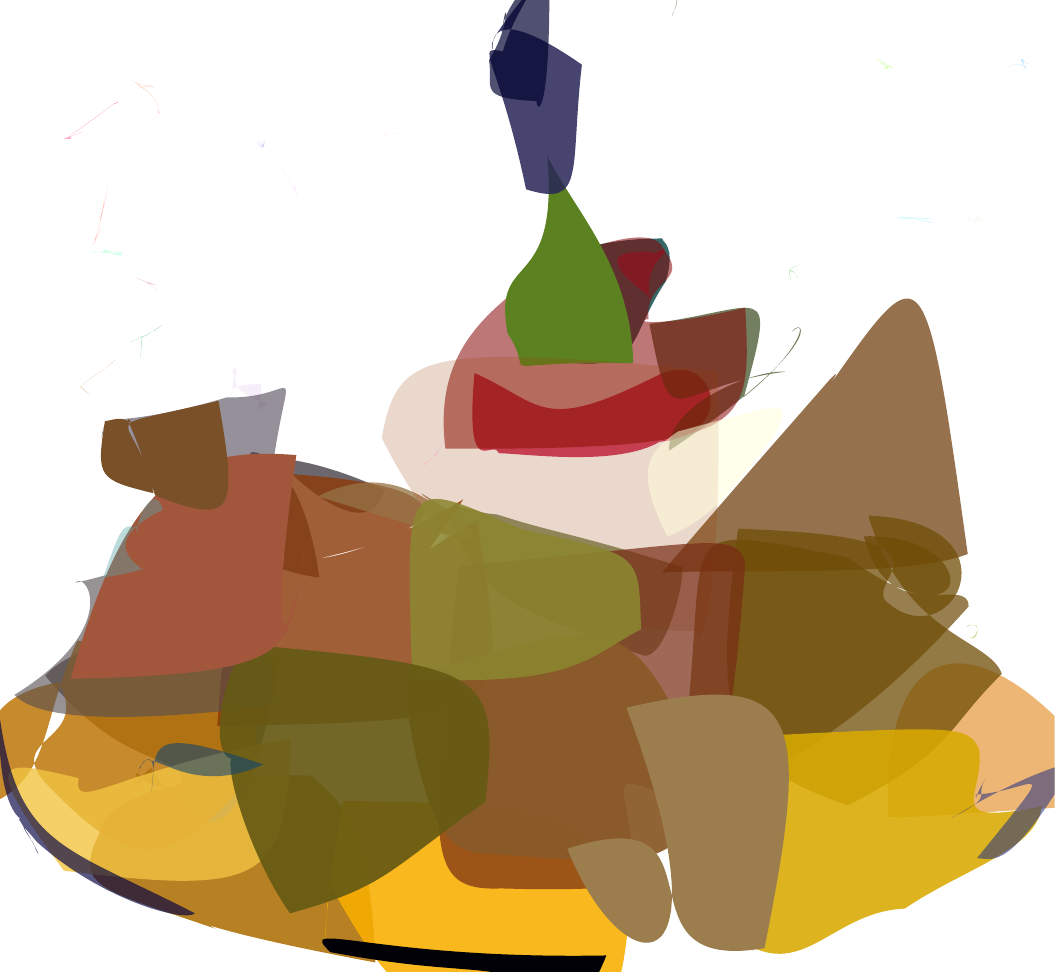} &
    \includegraphics[width=1.5cm, height=1.5cm]{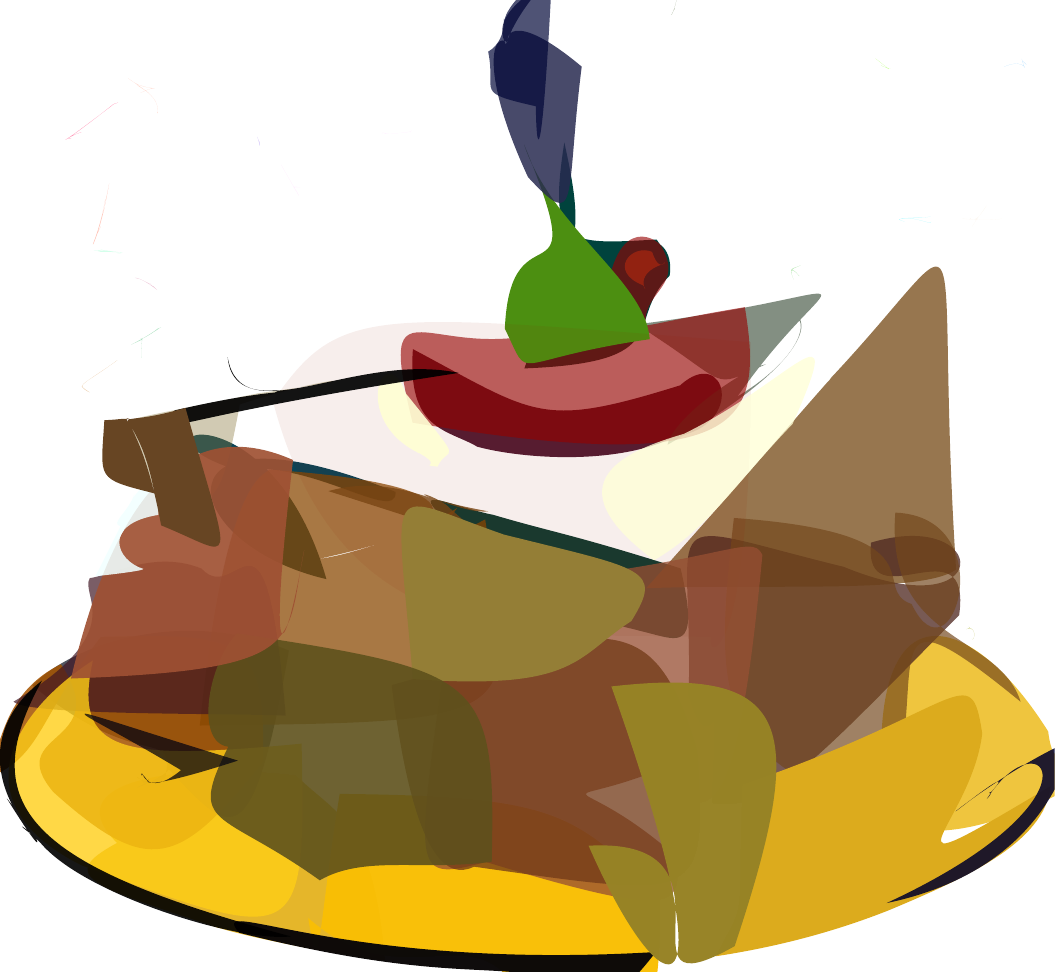} &
    \includegraphics[width=1.5cm, height=1.5cm]{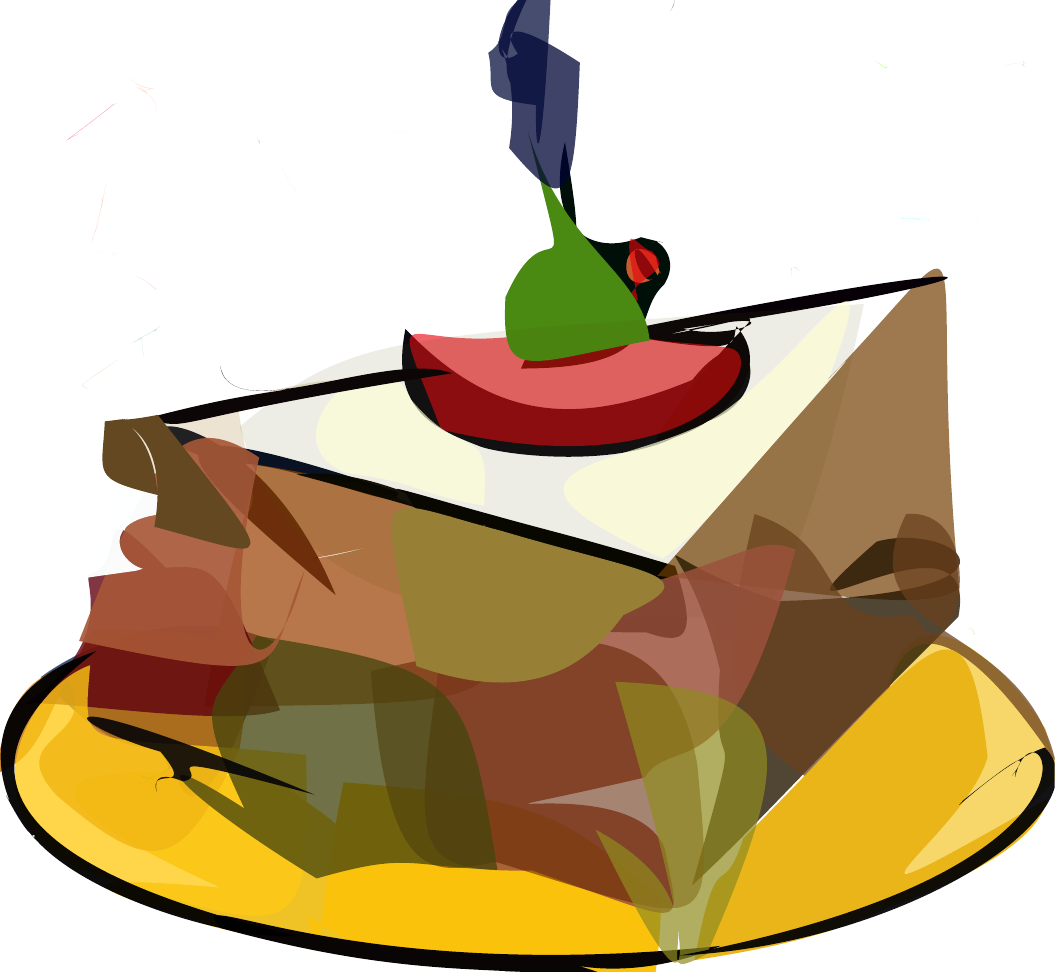} &
    \includegraphics[width=1.5cm, height=1.5cm]{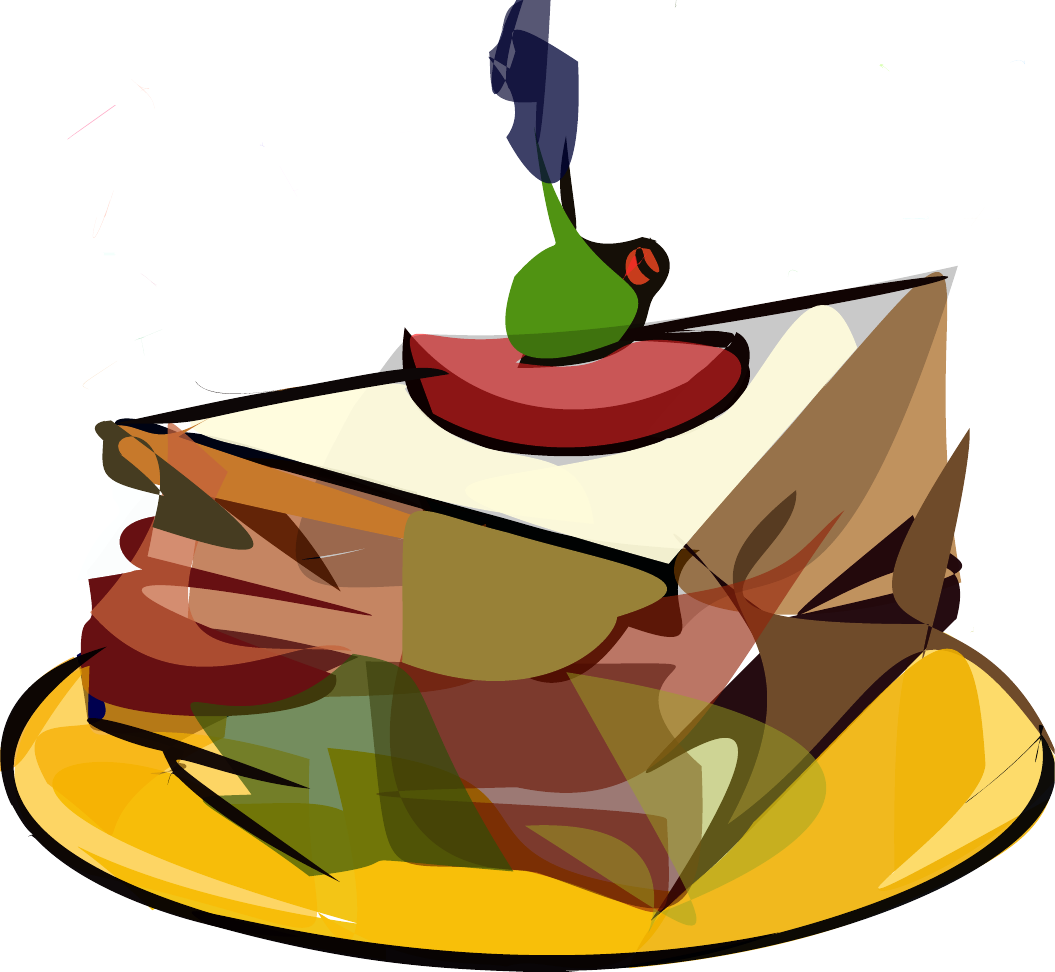} 
    \\
    \hline
    Original image&
    Step 1&
    Step 20&
    Step 50&
    Step 100&
    Step 200&
    Step 500
    \end{tabular}
\end{center}
\caption{Iterative vectorization using DiffVG method. Each generated image has 62 paths, the same number paths is used in the original vector image.}
\label{fig:diffvg}
\end{figure}

%The grounding work for machine learning-based vector image generation methods is the DiffVG~\cite{li2020diffvg}. 
The work suggesting DiffVG~\cite{li2020diffvg} is grounding for machine learning-based vector image generation methods as it implements a differentiable vector image rasterization function linking vector and raster domains.
%The authors implement a differentiable vector image rasterization function, thus linking vector and raster domains. 
A raster image can be vectorized with DiffVG by fitting a predefined number of randomly initialized B\'ezier curves to the target image, see Fig.~\ref{fig:diffvg}. 
Optimization can be performed by minimizing the $L_2$ loss between a raster and rasterized vector images  or a deep-learning-based perceptual loss~\cite{zhang2018unreasonable}.

% картинка про то, как кривые по итерациям все больше и больше повторяют растр

The authors also propose a simple variational autoencoder~\cite{kingma2013auto} for vectorizing MNIST digits images~\cite{deng2012mnist}. 
The encoder convolves the input raster image into a latent space, a vector image is synthesized from the embedding, then the input and synthesized images are compared using the proposed rasterization function.
The resulting images are inaccurate because each character is represented by several curves that look like the artist's strokes. 
The results do not exactly match the original images of such a simple dataset as MNIST.

One of this method features is also worth mentioning. 
The running time of the rasterization algorithm significantly depends on the size of the output image. However, output size reduction result in the loss of important image details. This problem arises due to the nature of vector images. 
%Although vector images are claimed to be resolution-independent, their rendering to small resolutions lose these detailsthey images containing a lot of detail when .
By analogy with raster images size decrease, that leads to the loss of image details, the same effect occurs when the number of paths is reduced in vector images. 
% which raster image size is decrease it lose details, 
% then number of paths is reduced 
%At the same time, when the output sizes are reduced, important details of the image may be lost. 
%This problem arises due to the nature of vector images, which, although they are proclaimed as image resolution-independent, but at the same time images containing a lot of detail when rendering to a small resolution lose these details.

\subsection{Im2Vec}
The Im2Vec~\cite{reddy2021im2vec} paper offers a model for image vectorization and interpolation. 
Its architecture is based on the variational auto-encoder (VAE)~\cite{kingma2019introduction} proposed in DiffVG. 
The model maps an input raster image into a latent space sequentially generating a similar vector image. 
While training, to compare the generated vector image with the input raster image, the vector image is rasterized using a differentiable rasterizer. 

In the paper, the authors propose a new way of generating closed shapes.
Initially, a circle is sampled for each shape, and then, based on the latent vectors of the shapes generated with LSTM, the circle deforms. 

Since the difference between the target and output images is significant at the beginning of training, the authors suggest using multi-resolution loss. 
The paper proposes to rasterize images in different resolutions, thus building a pyramid of images, with the loss function for each layer being calculated.
% The authors used an image pyramid as a representation and aggregated loss at each pyramid level. 
The total multi-resolution loss optimized is: \[
\mathbb{E}_{I\sim D} \sum_{l=1}^{L} \|pyr_l (I) - O_l \|^2,
\] 
where $L$ is the number of pyramid levels, $pyr_l(I)$ is the $l$-th pyramid level, $O_l$ is output rasterized at the corresponding spatial resolution, and $D$ is the training dataset.
%  We tried to reproduce the results of the work, but the official implementation is apparently incorrect. 
Unfortunately, using the official implementation, the results can hardly be reproduced.
% Initially, the authors indicated very little information about the installation of libraries and their dependencies. After solving these problems, we launched the model on the smiley dataset proposed by the authors with the specified and other hyperparameters.
During the first $700$ epochs of training on the emoji dataset, which was collected and published by the authors, the generation result has no changes and consists of a single point in the center of the image. 
We have also noticed that the shapes colors are fixed in the implementation, i.e. no separate color generation can be performed, so we have added a separate LSTM model for predicting the colors of figures. 
However, resolving issues to their implementation, the authors themselves point out that color prediction causes instability during generation.

Thus, this work cannot be considered as a universal model for vectorizing any image, because: 1) it must be pretrained on a large number of vector images, which are hard to obtain; 2) the found shortcomings are likely to lead to learning instability providing resulting images of poor quality.

\subsection{LIVE}
In paper~\cite{ma2022towards}, LIVE vectorizer was introduced. 
LIVE is a logical continuation of the iterative method proposed in DiffVG. 
However, LIVE does not operate with all shapes simultaneously from the first iteration. Instead, it gradually adds one or more shapes to the canvas layer by layer and then performs an optimization step.

Unlike DiffVG, LIVE operates only with closed shapes consisting of cubic B\'ezier curves. There is an issue that some of them may become self-interacted during optimization, which results in undesirable artifacts and incorrect topology. Although additional paths could cover artifacts, it would complicate the resulting SVG making it impossible to effectively investigate the underlying topological data. The authors discovered that a self-intersecting path always intersects the lines of its control points, and vice versa, assuming that all of the B\'ezier curves are of the third order. 
Therefore, the authors introduce a new loss function (Xing-loss) designed to solve this self-intersection problem. 
The fundamental idea is to only optimize the case when the angle of the curve is 180\textdegree  degrees. In other words, the authors urge the angle between the first and last control points connections to be greater than 180\textdegree  in a cubic B\'ezier path.
This loss acts as a regularizer on self-intersection and its formula is:
\[
\mathcal{L}_{Xing} = D_1 (ReLU(-D_2))  (1 - D_1)(ReLU(D_2))
\]
where $D_1$ is a characteristic of the angle between two segments of a cubic B\'ezier path, and $D_2 = \sin{\alpha}$ {{---}} value of that angle.

To make each path responsible only for a single feature of the image, the authors introduce Unsigned Distance guided Focal loss (UDF loss) as well; it treats each pixel differently depending on how close it is to the shape contour.
According to intuition, the UDF loss amplifies differences near the contour and suppresses differences in other areas {{---}} LIVE weighs an $L_2$ reconstruction loss by distance to the nearest path. 
By doing this, LIVE defends against the mean color problem caused by MSE and keeps accurate color reconstruction:
\[
\mathcal{L}_{UDF} = \frac{1}{3} \sum_{i=1}^{w\times h} d'_i \sum_{c=1}^{3} (I_{i,c} - \hat{I_{i,c}})^2,
\]
where $I$ is the target image, $\hat{I}$ is the rendering, $c$ indexes RGB channels in $I$, $d'_i$ is the unsigned distance between pixel $i$, and the nearest path boundary, and $w$, $h$ are width and height of the image. 

\begin{figure}[h]
\begin{center}
    \begin{tabular}{
    >{\centering\arraybackslash}m{1.6cm} |
    >{\centering\arraybackslash}m{1.6cm}
    >{\centering\arraybackslash}m{1.6cm}
    >{\centering\arraybackslash}m{1.6cm}
    >{\centering\arraybackslash}m{1.6cm}
    >{\centering\arraybackslash}m{1.6cm}
    >{\centering\arraybackslash}m{1.6cm}
    }
    \includegraphics[width=1.6cm, height=0.95cm]{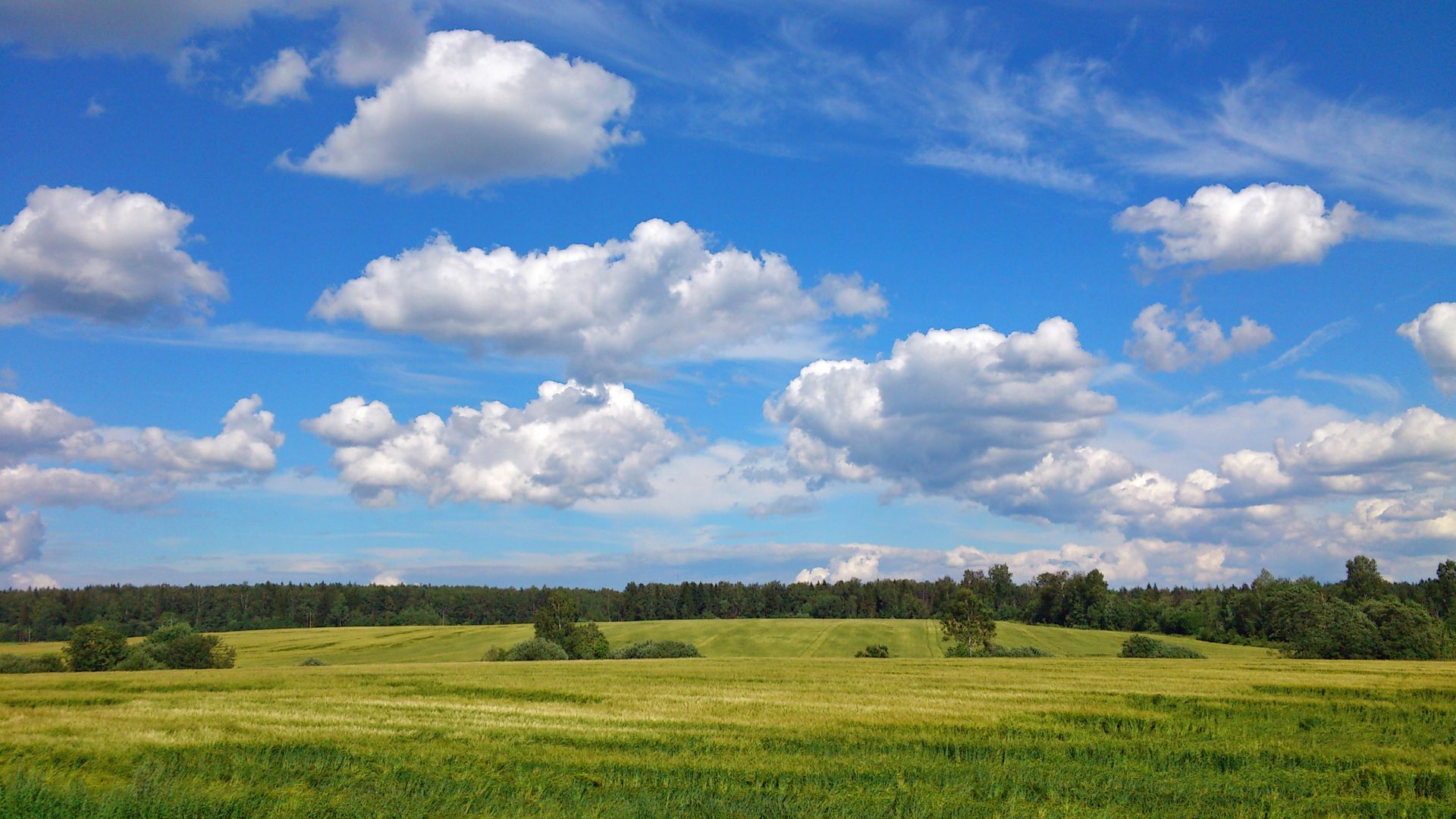} &
    \includegraphics[width=1.6cm, height=1.2cm]{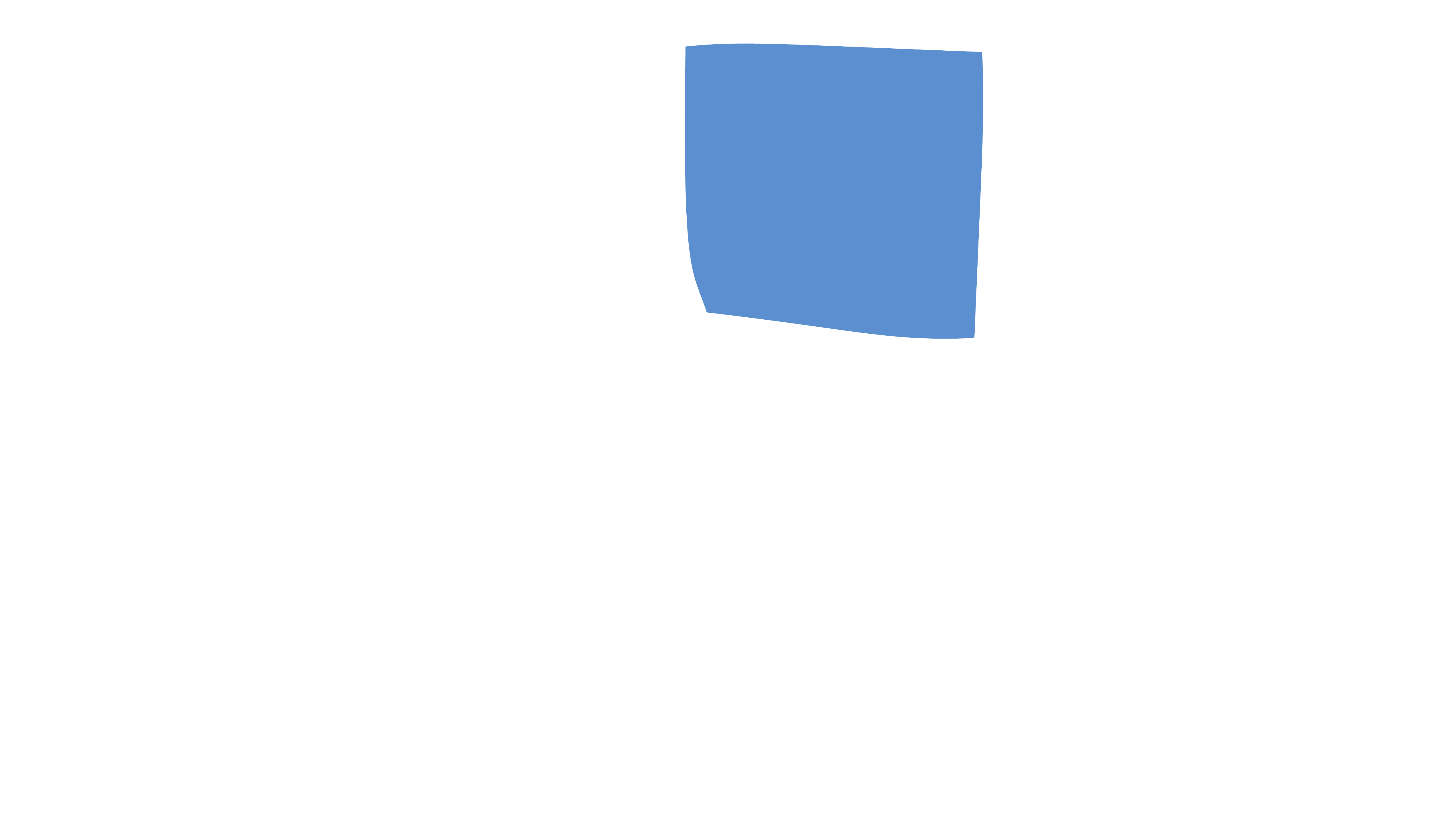} &
    \includegraphics[width=1.6cm, height=1.2cm]{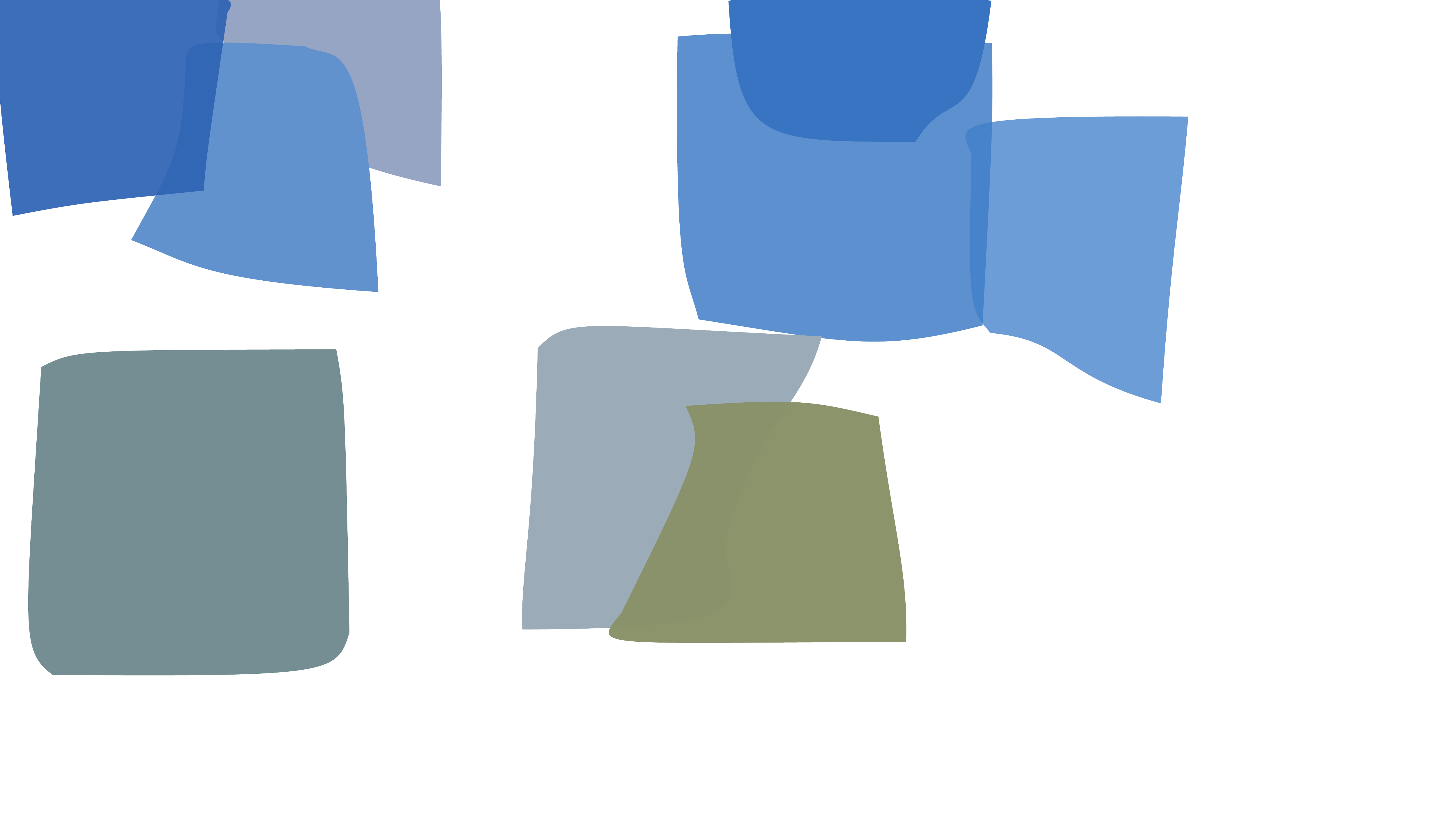} &
    \includegraphics[width=1.6cm, height=1.2cm]{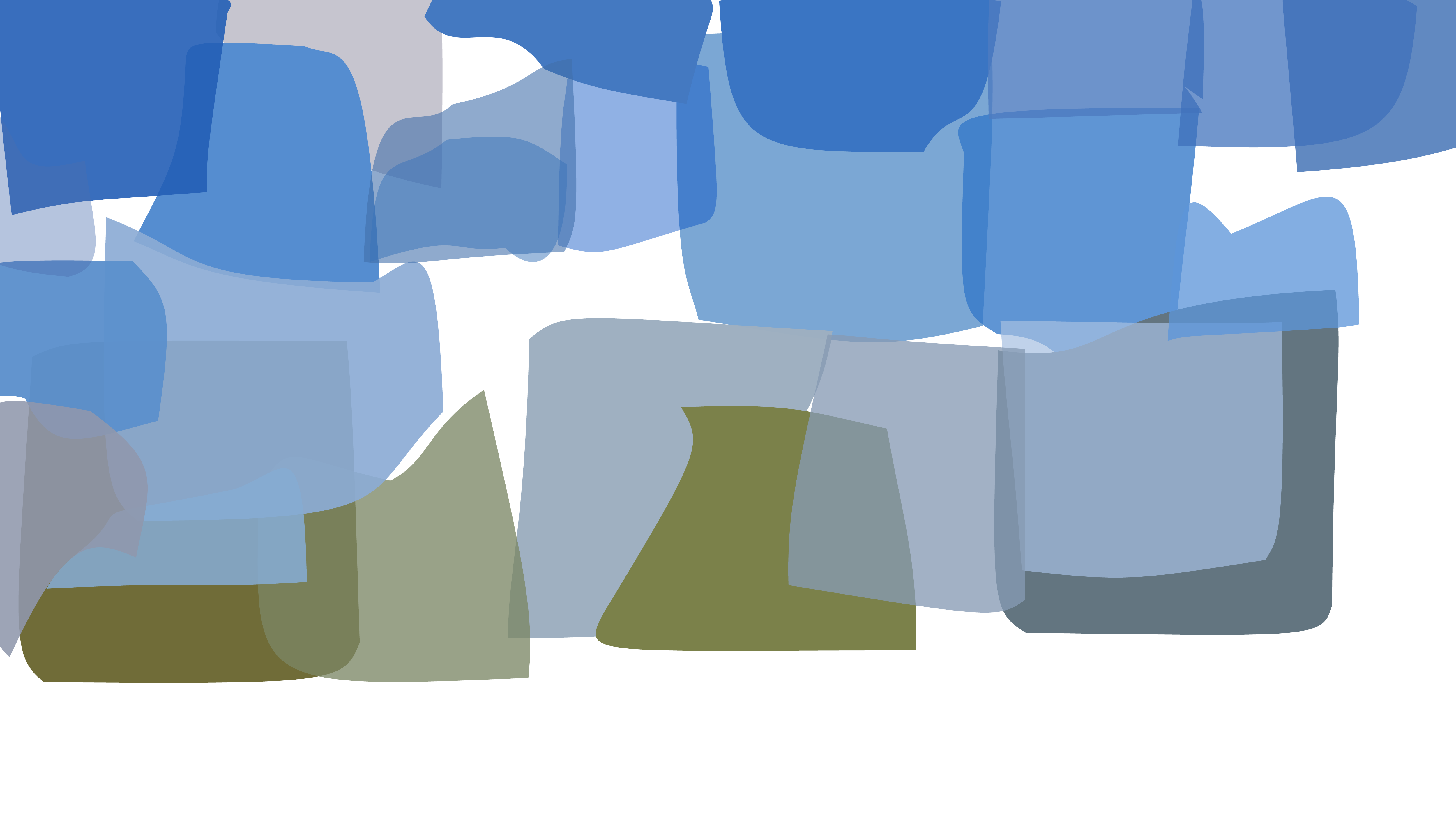} &
    \includegraphics[width=1.6cm, height=1.2cm]{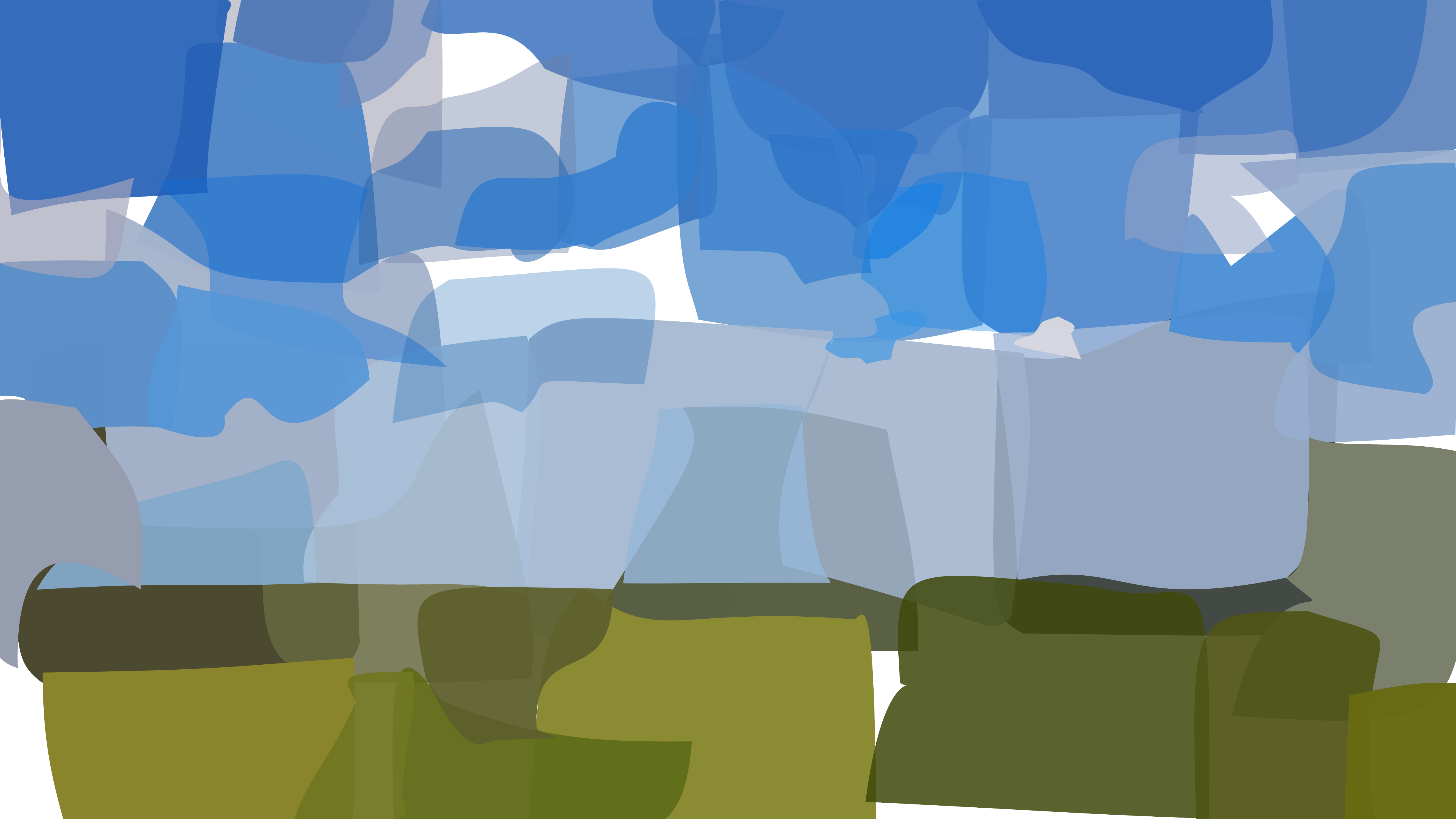} &
    \includegraphics[width=1.6cm, height=1.2cm]{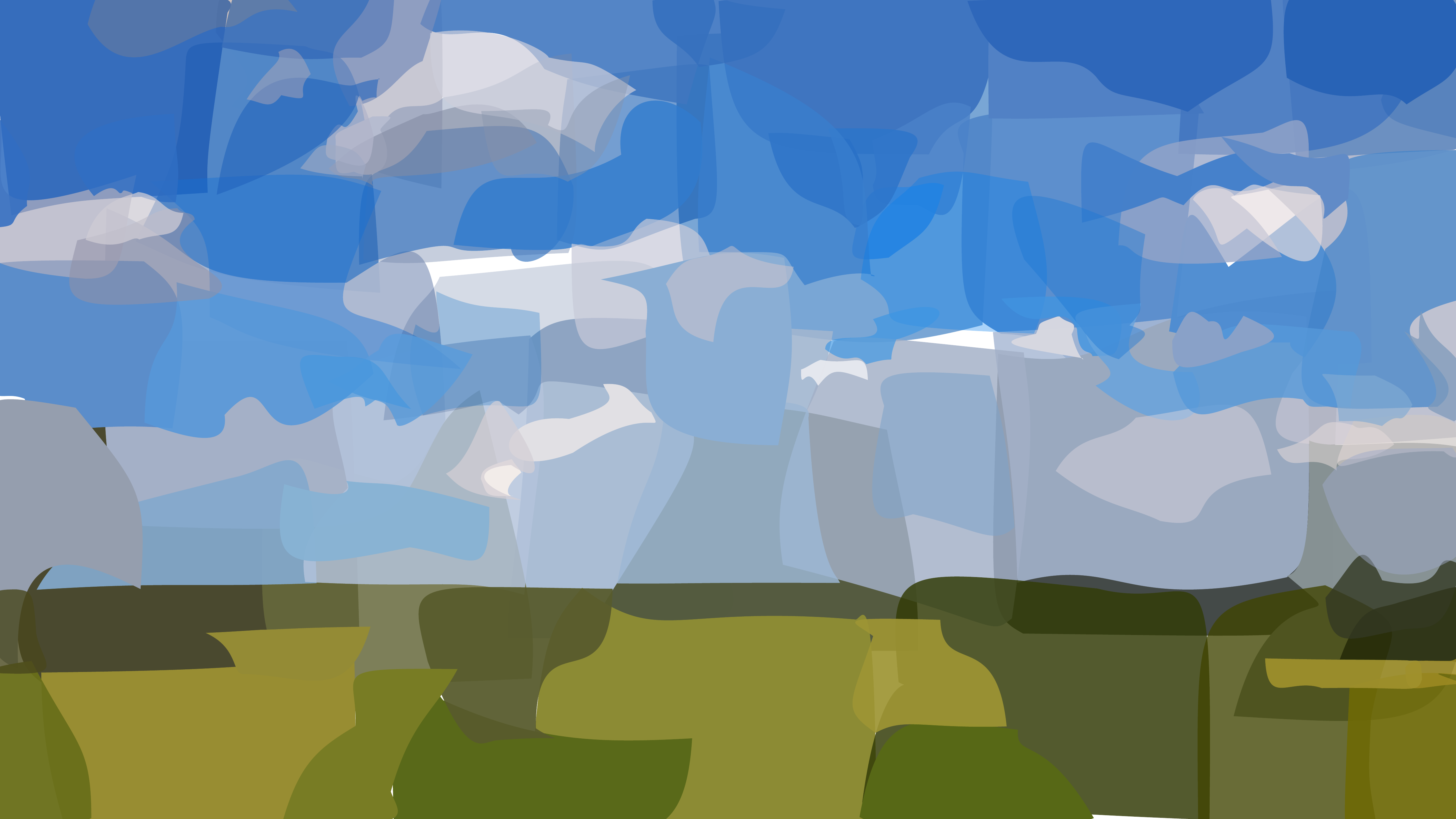} &
    \includegraphics[width=1.6cm, height=1.2cm]{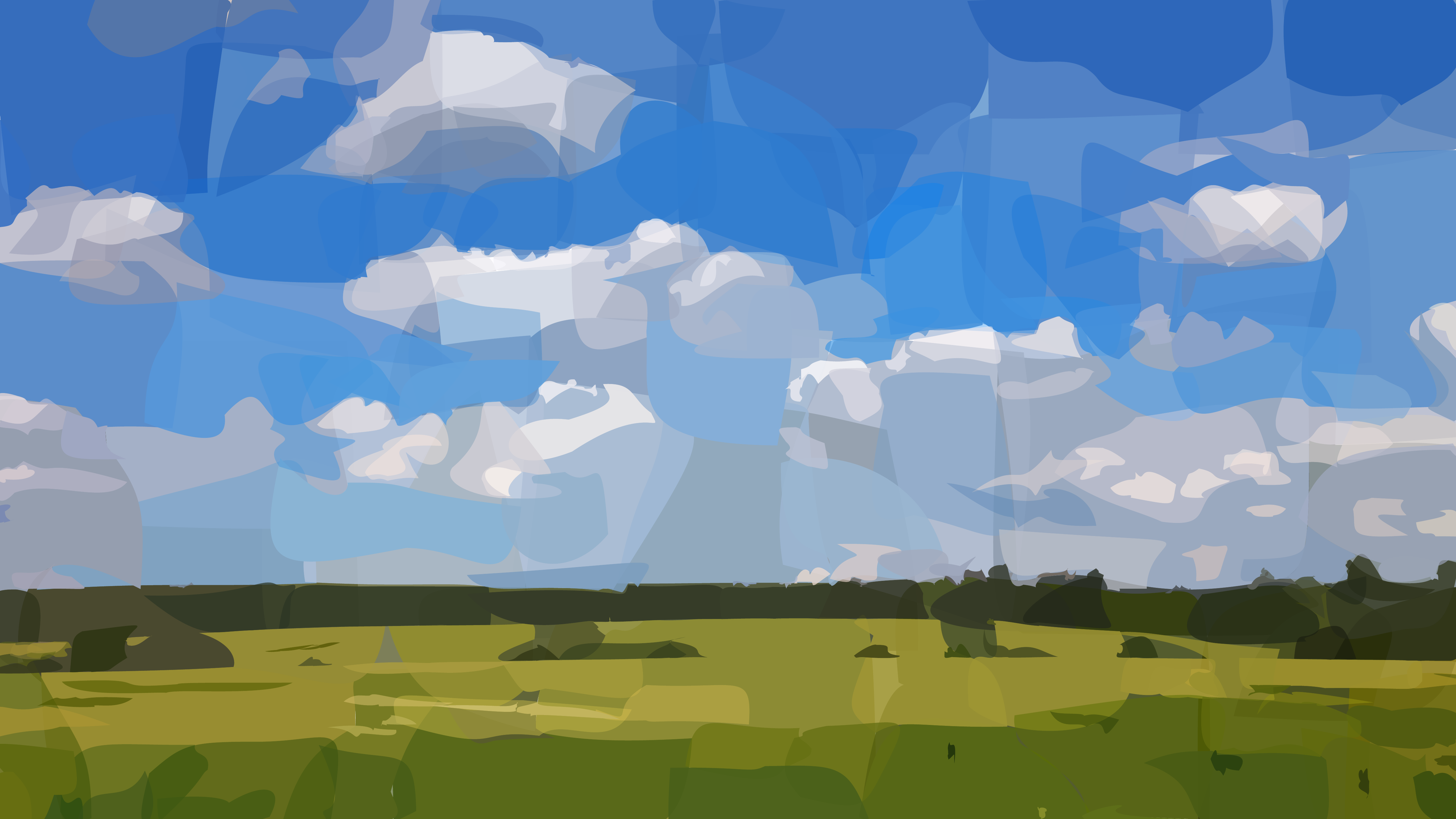} 
    \\
    \hline
    Original image&
    1 figure (1~layer)&
    9 figures (3~layers)&
    26 figures (5 layers)&
    56 figures (8 layers)&
    106 figures (13 layers)&
    256 figures (28 layers)
    \end{tabular}
\end{center}
\caption{Iterative vectorization using LIVE method.}
\label{fig:LIVE_steps}
\end{figure}

LIVE produces relatively clean SVGs by initializing paths in stages, localized to poorly reconstructed, high-loss regions. LIVE's main advantage is its ability to reconstruct an image with a user-defined amount of paths, significantly reducing the SVG file size compared to other methods. However, it takes much time to vectorize an image even on GPU, thus, this method is hardly applicable in practice for complex images with a great optimal number of paths.
See the iterative process in Fig.~\ref{fig:LIVE_steps}.

% The method generates an SVG image that fits the arbitrary raster image layer by layer by adding and optimizing new closed B\'ezier curves. 

% Было в тексте:
% Although a variety of shape primitives can be added to an SVG, the authors consider the parametric closed B\'ezier curves as the most significant shape primitive, see Fig.~\ref{fig:LIVE_steps}.

\subsection{ClipGen}
The ClipGen paper~\cite{shen2021clipgen} proposes a method based on deep learning for automatically vectorizing the clipart of man-made objects. 
The suggested approach needs a raster clipart image and relevant object category (for instance, airplanes). 
It sequentially creates new layers, each formed by a new closed path that is filled with a single color. 
All layers are combined to create a vector clipart that fits the desired category to produce the resulting image. 
The suggested method is built on an iterative generative model that chooses whether to keep synthesizing new layers and defines their geometry and appearance. 
For training their generative model, they developed a joint loss function that includes shape similarity, symmetry, and local curve smoothness losses, as well as vector graphics rendering accuracy loss for synthesizing a human-recognizable clipart. 
However, ClipGen only works with a predefined number of categories, therefore, it cannot process arbitrary images.  
%so it has to be finetuned before usage.

\subsection{Mang2Vec}
The authors of the Mang2Vec ~\cite{su2021vectorization} paper suggest the first method for vectorizing raster mangas by using Deep Reinforcement Learning.  They develop an agent that is trained to generate the best possible sequence of stroke lines while being constrained to match the target manga visual features. The control parameters for the strokes are then collected and converted to the vector format. They also propose a reward to produce accurate strokes and a pruning method to avoid errors and redundant strokes. 
Mang2Vec works only with black and white manga and cannot be used with colored images. 
%Also it generates only stroke lines.

\begin{figure}[h]
\begin{center}
    \begin{tabular}{
    >{\centering\arraybackslash}m{1.6cm} |
    >{\centering\arraybackslash}m{1.6cm}
    >{\centering\arraybackslash}m{1.6cm}
    >{\centering\arraybackslash}m{1.6cm}
    >{\centering\arraybackslash}m{1.6cm}
    >{\centering\arraybackslash}m{1.6cm}
    >{\centering\arraybackslash}m{1.6cm}
    }
    \includegraphics[width=1.6cm, height=1.2cm]{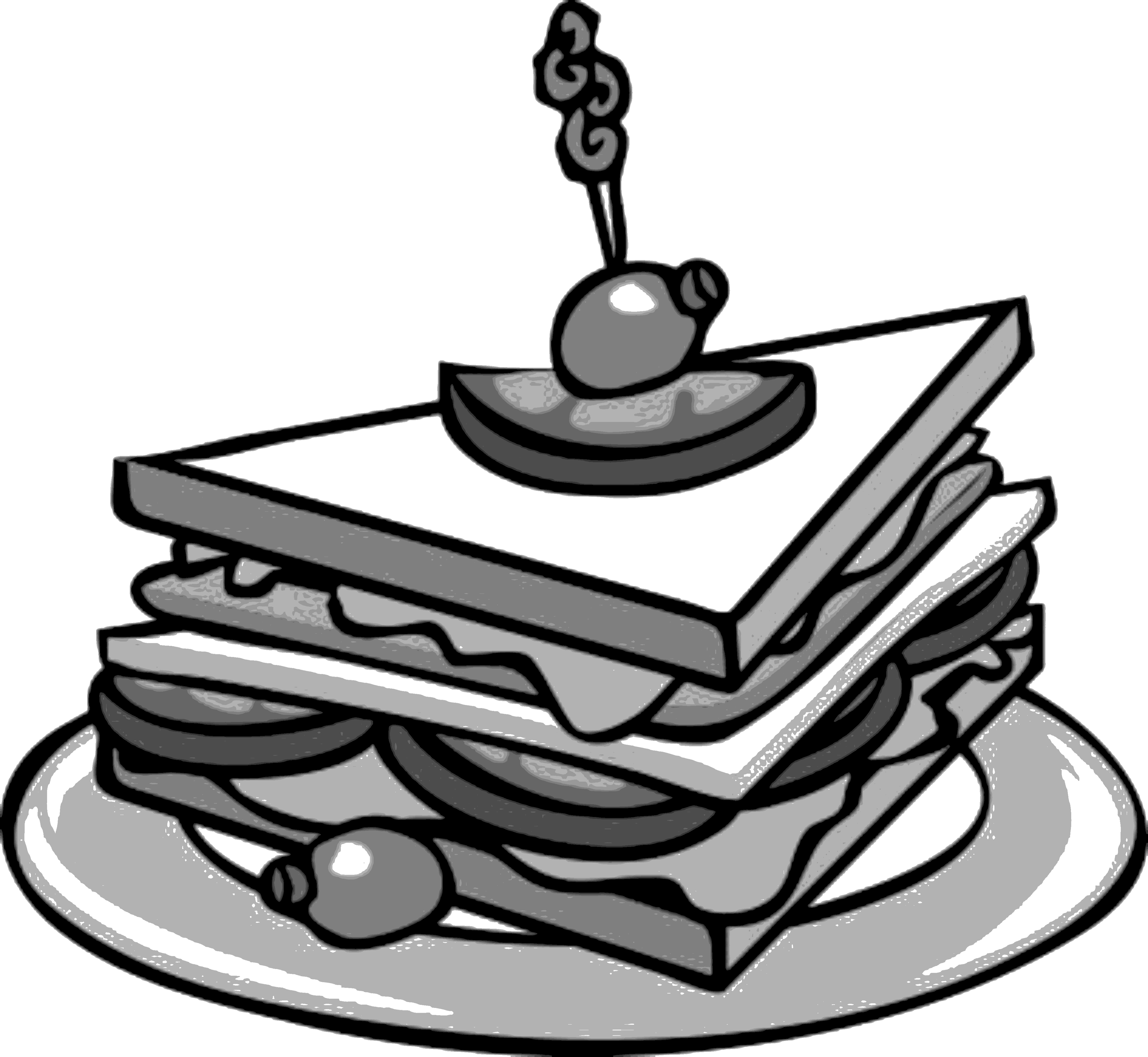} &
    \includegraphics[width=1.6cm, height=1.2cm]{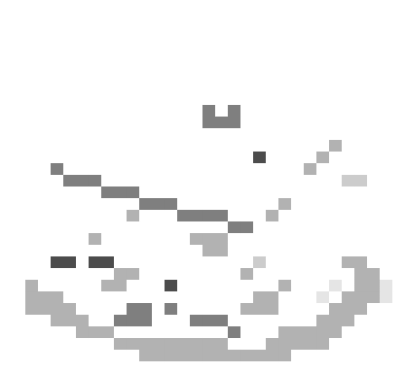} &
    \includegraphics[width=1.6cm, height=1.2cm]{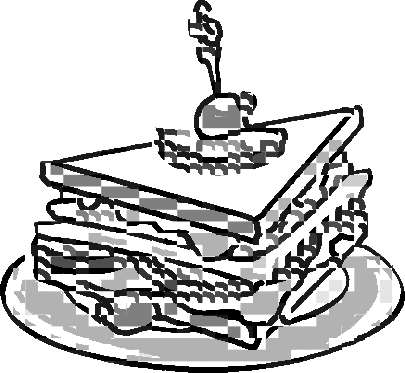} &
    \includegraphics[width=1.6cm, height=1.2cm]{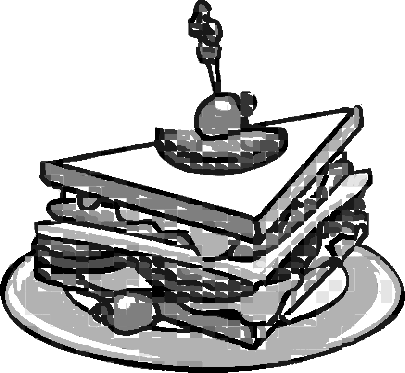} &
    \includegraphics[width=1.6cm, height=1.2cm]{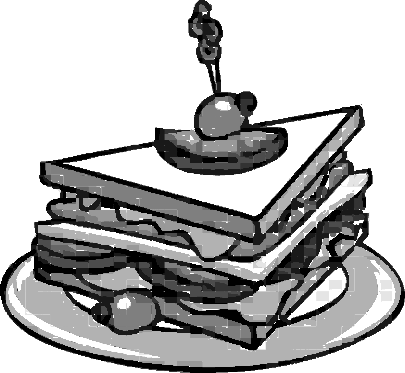} &
    \includegraphics[width=1.6cm, height=1.2cm]{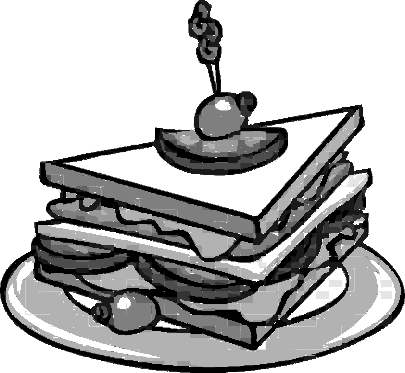} &
    \includegraphics[width=1.6cm, height=1.2cm]{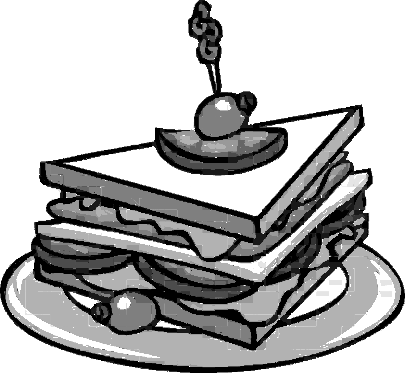}
    \\
    \hline
    Original image&
    Init step 0&
    Step 1&
    Step 2&
    Step 3&
    Step 4&
    Step 5
    \end{tabular}
\end{center}
\caption{Vectorization using Mang2Vec method.}
\label{fig:mang2vec_steps}
\end{figure}

\subsection{Deep Vectorization of Technical Drawings}
The paper~\cite{egiazarian2020deep} proposes a technical line drawings vectorization method (DVoTD), for example, for drawings of floor plans. The authors convert a technical raster drawing, which is cleared of text, into a set of line segments and quadratic B\'ezier curves that are specified by control points and width. They preprocess the input image by eliminating noise, modifying contrast, and adding missing pixels. Then, they divide the image into patches and calculate the starting primitive parameters for each patch. To do this, each patch is encoded with a ResNet-based feature extractor and decoded as feature embeddings of the primitives using a sequence of transformer blocks. To train the network for primitive extraction, the following loss function is proposed: 
\[
L(p, \hat{p}, \theta, \hat{\theta}) = \frac{1}{n_{prim}} \sum_{k=1}^{n_prim} (L_{cls} (p_k, \hat{p_k}) + L_{loc} (\theta_k, \hat{\theta_k}),
\]
where 
\[
L_{cls}(p_k, \hat{p_k}) = - \hat{p_k} \log p_k - (1-\hat{p_k}) \log (1 - p_k),
\]
\[
L_{loc} (\theta_k, \hat{\theta_k}) = (1- \lambda) \| \theta_k - \hat{\theta_k} \|_1 + \lambda \| \theta_k \hat{\theta_k}\|_2^{2},
\]
$\hat{p}$ {{---}} the target confidence vector (is all ones, with zeros in the end that indicate placeholder primitives, all target parameters $\hat{\theta_k}$ of which are set to zero).

The approximated primitives improve by aligning to the cleaned raster. The improved predictions from all patches are combined. 

\section{Online Vectorization Methods}
There are plenty of websites that can vectorize any raster image. 
Existing online methods can be free to use (svgstorm.com, www.visioncortex.org/vtracer, vectorization.eu) or proprietary (vectorizer.io).

Common options provided by these methods is the selection of vector graphics output file format (SVG, EPS, PDF), color palette and the number of colors used. 
Some services allow choosing the quality of image detail (Low, Medium, High), the type of shapes used (Curve Fitting - pixel/polygon/curve), background removal and many other actions and parameters.
These options affect the processing speed, the visual result and the number of shapes.
The generation speed highly depends on the resolution of the input raster image and its details complexity. On average, the processing time of one image is $10$ seconds.

Even though they are easy to use, a lot of parameters should be fixed by a user. It should be mentioned that the resulting image quality and its size hardly depend on the input image quality and resolution. Low quality results in producing images with a lot more paths that are actually needed leading to noticeable artifacts.
% , as can be seen in Fig.~\ref{fig:online}. 

% VTracer
One of the popular online services for vectorization is VTracer~\cite{vtracer}, which provides many options for a user. 
According to its documentation, firstly the method clusters the input image by hierarchical clustering and each of the output clusters traces into vector. 
After converting pixels into staircase-like paths, the method then simplifies the paths into polygons and in the end smoothens and approximate them with a curve-fitter. 
%VTracer has a web app, and a user can adjust the following parameters: number of channels (black and white or RGB), the way clustering is performed (clustering shapes disjoint with others or shapes are stacked on top of each other), filter speckle (algorithm discards patches that are smaller than X px in size), color precision (number of significant bits to use in a RGB channel), gradient step, curve fitting type (pixel, polygon or curve), minimum angle ...
% The input image is first clustered by hierarchical clustering and each of the output clusters are traced into vector. 
% It first converts pixels into paths, then simplifies the path into polygon and after that smoothens the polygon and approximate it with a curve-fitter. 

\begin{figure}[h]
\begin{center}
    \begin{tabular}{
    >{\centering\arraybackslash}m{1.9cm}
    >{\centering\arraybackslash}m{1.9cm}
    >{\centering\arraybackslash}m{1.9cm}
    >{\centering\arraybackslash}m{1.9cm}
    }
    \includegraphics[width=1.9cm]{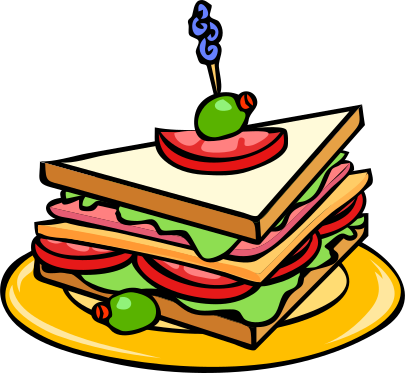} &
    \includegraphics[width=1.9cm]{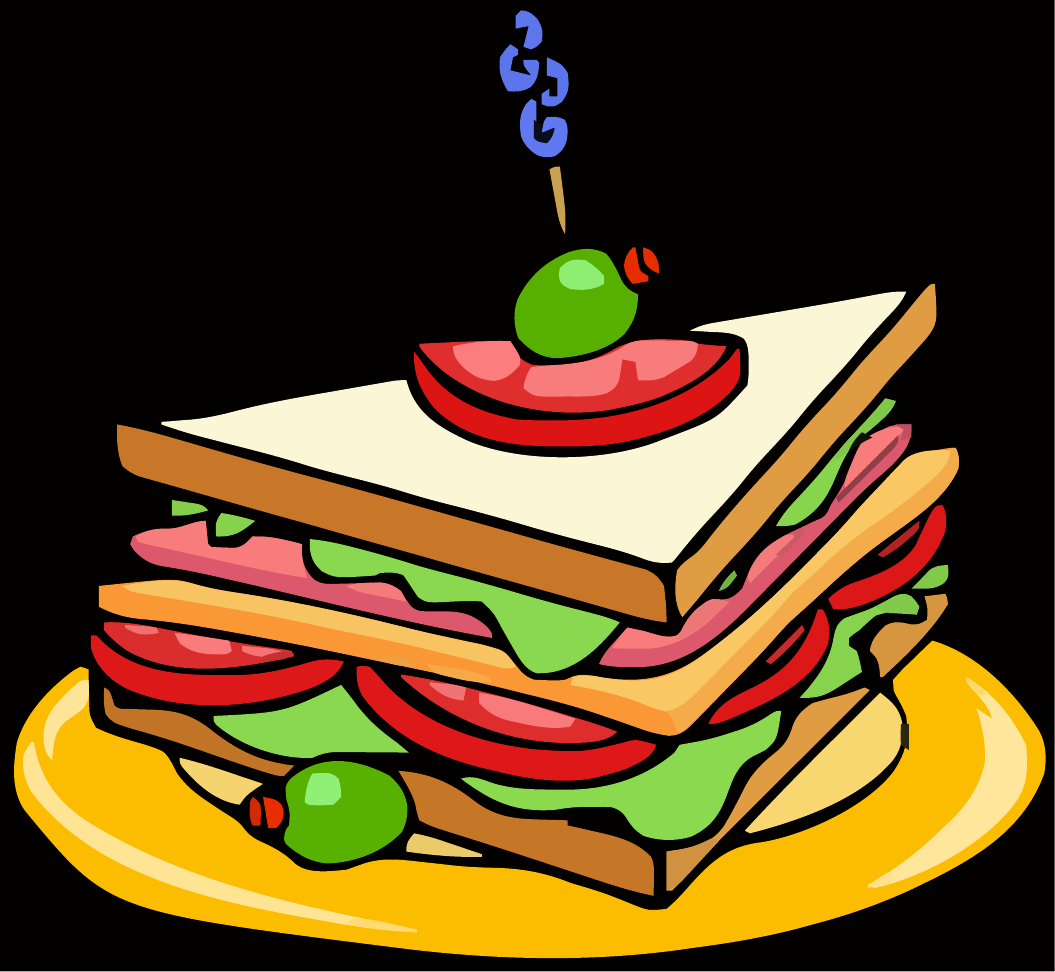} &
    \includegraphics[width=1.9cm]{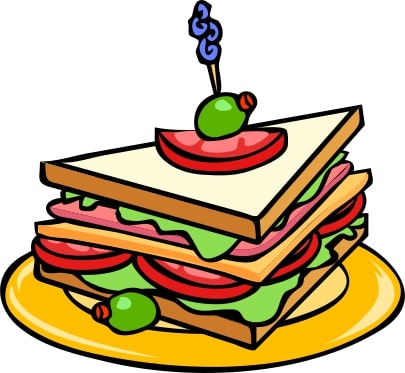} &
    \includegraphics[width=1.9cm]{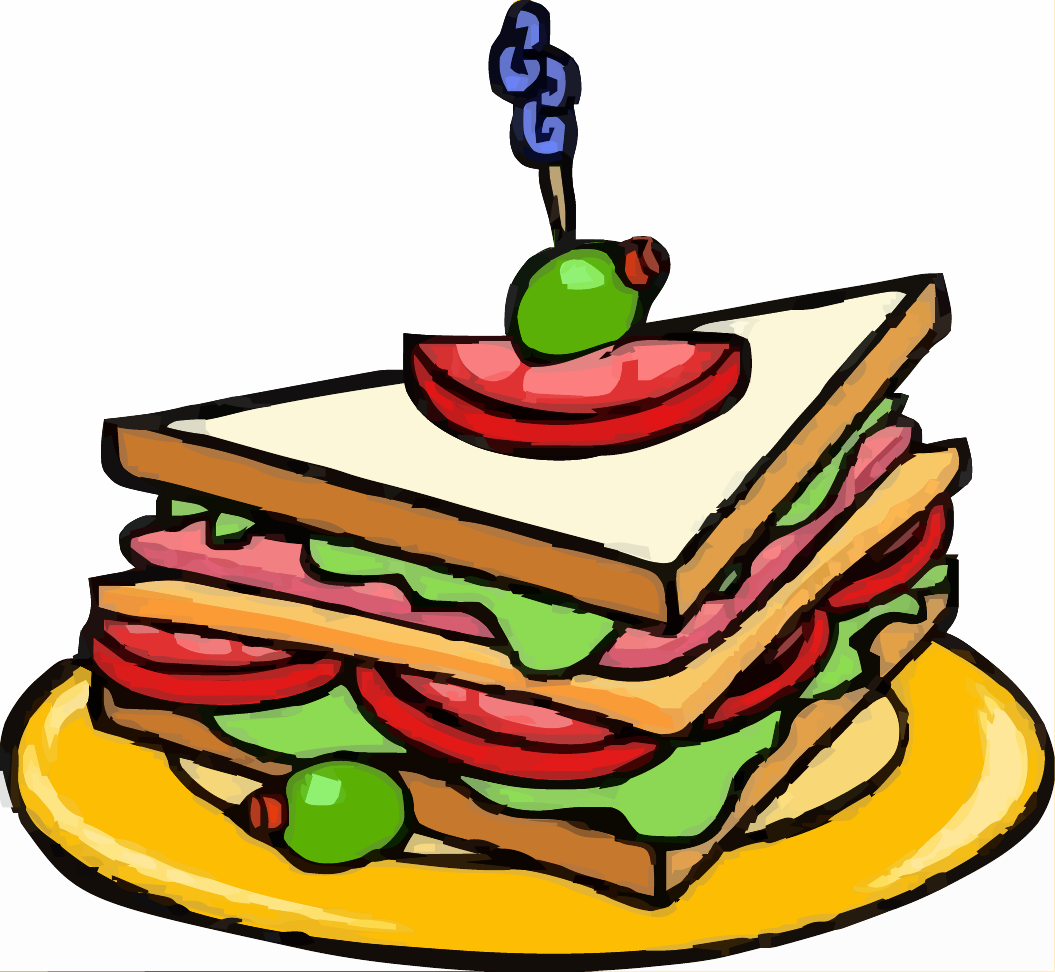}
    \\
    \hline
    Original PNG image&
    PNG vectorized with black background (81 paths)&
    Original JPG image&
    JPG vectorized (720 paths)
    \end{tabular}
\end{center}
\caption{VTracer vectorization and its issues with the black background color and inaccurate vectorization of low-quality images.}
\label{fig:vtracer}
\end{figure}

% алгоритм работает хорошо не со всеми форматами (например,у png черный фон), 
We have come across the following drawbacks of this service.
Firstly, VTracer does not work well with all image formats, for example, it produces a black background instead of a transparent one while processing PNGs and there are no options to change this behaviour.
Secondly, VTracer does not handle low-quality images well, creating many unnecessary inaccurate shapes. 
In Fig.~\ref{fig:vtracer}, we show an example of the black background appearance for a PNG high-quality image and the result for the same image converted to JPG having low quality.

% \begin{figure}[h]
% \begin{center}
%     \begin{tabular}{
%     >{\centering\arraybackslash}m{1.6cm} |
%     >{\centering\arraybackslash}m{1.6cm}
%     >{\centering\arraybackslash}m{1.6cm}
%     >{\centering\arraybackslash}m{1.6cm}
%     >{\centering\arraybackslash}m{1.6cm}
%     >{\centering\arraybackslash}m{1.6cm}
%     >{\centering\arraybackslash}m{1.6cm}
%     }
%     \includegraphics[width=1.6cm, height=1.2cm]{images/original/burger2.pdf} &
%     \includegraphics[width=1.6cm, height=1.2cm]{images/svgstorm/burger/burger2_default_med.pdf} &
%     \includegraphics[width=1.6cm, height=1.2cm]{images/svgstorm/burger/burger2_4_med.pdf} &
%     \includegraphics[width=1.6cm, height=1.2cm]{svgstorm/burger/burger2_8_med.pdf} &
%     \includegraphics[width=1.6cm, height=1.2cm]{images/svgstorm/burger/burger2_16_med.pdf} &
%     \includegraphics[width=1.6cm, height=1.2cm]{images/svgstorm/burger/burger2_32_med.pdf} 
%     \\
%     \hline
%     Original image (62 paths)&
%     Default (184 paths)&
%     4 colors (132 paths)&
%     8 colors (174 paths)&
%     16 colors (262 paths)&
%     32 colors (338 paths)
%     \end{tabular}
% \end{center}
% \caption{Online vectorization on svgstorm.com.}
% \label{fig:online}
% \end{figure}

\section{Comparison}

\begin{figure}[h]
\begin{center}
\begin{tabular}{
>{\centering\arraybackslash}m{1.5cm} |
>{\centering\arraybackslash}m{1.65cm}
>{\centering\arraybackslash}m{1.65cm}
>{\centering\arraybackslash}m{1.65cm}
>{\centering\arraybackslash}m{1.65cm}
>{\centering\arraybackslash}m{1.65cm}
>{\centering\arraybackslash}m{1.65cm}
}
\toprule
Image Description&
Original&
Mang2Vec&
DVoTD&
DiffVG (closed)&
DiffVG (unclosed)&
LIVE \\
\midrule
% 1
Simple Vector
&
\includegraphics[width=1.65cm]{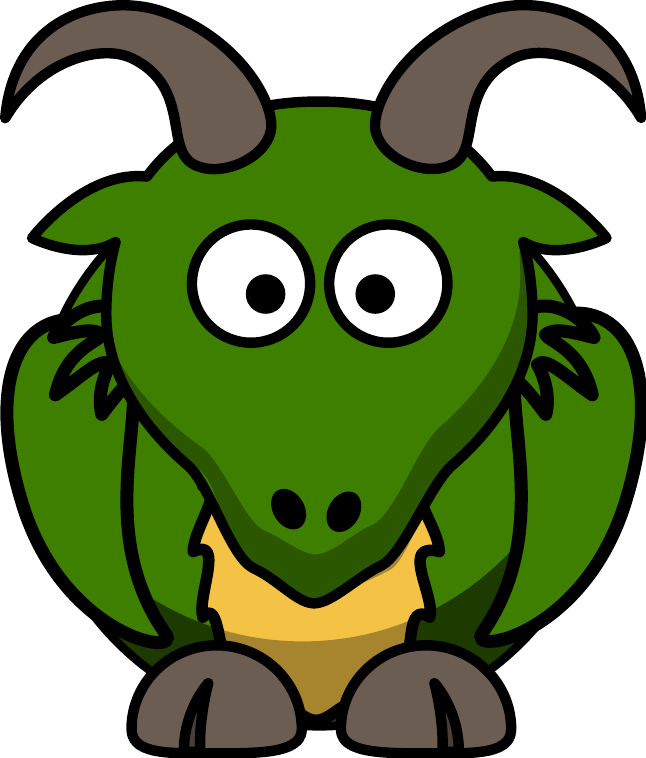}
&
\includegraphics[width=1.65cm]{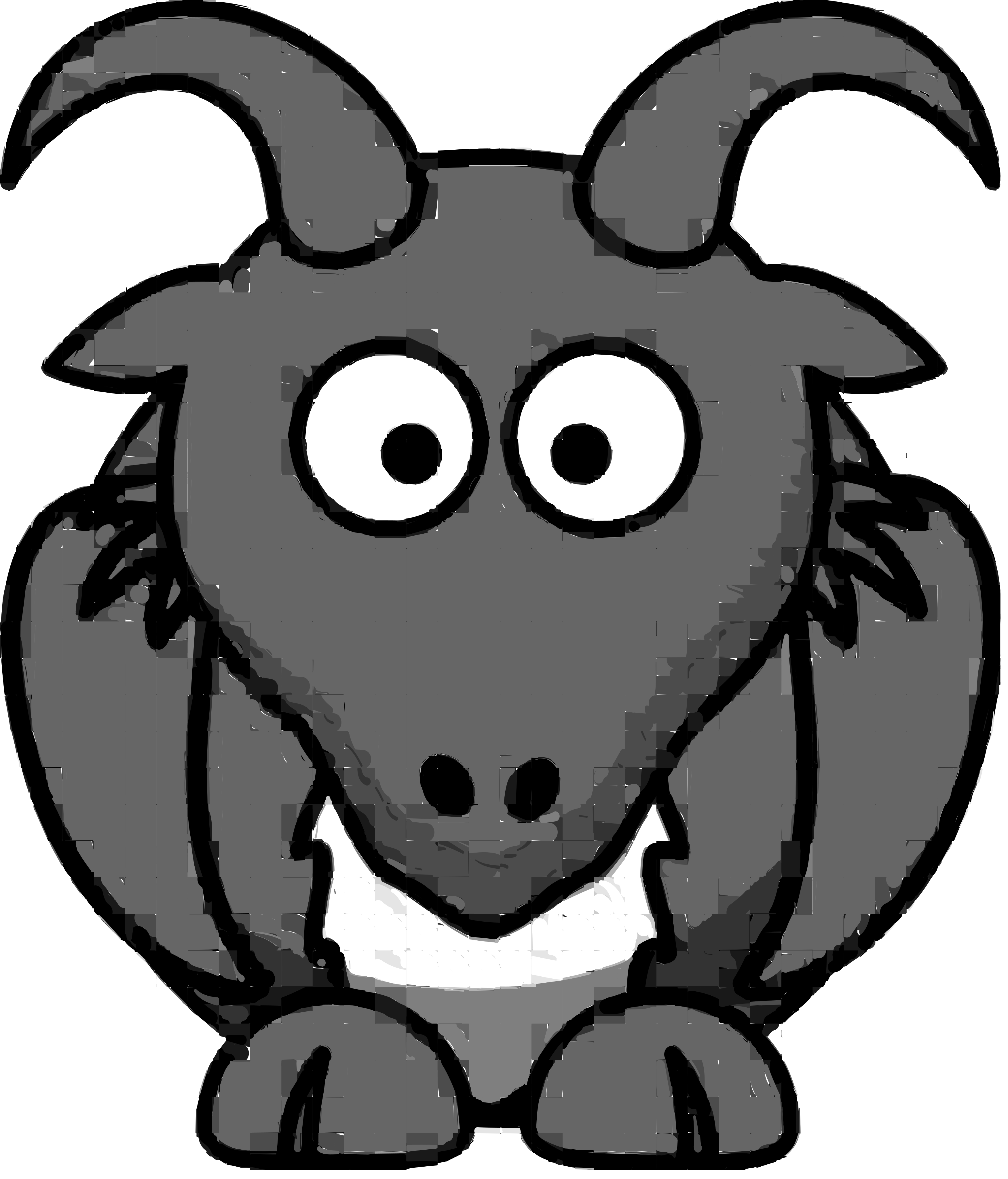}
&
\includegraphics[width=1.65cm]{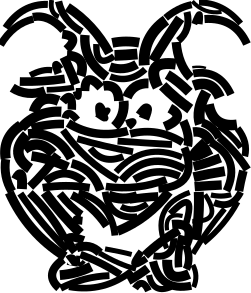}
&
\includegraphics[width=1.65cm]{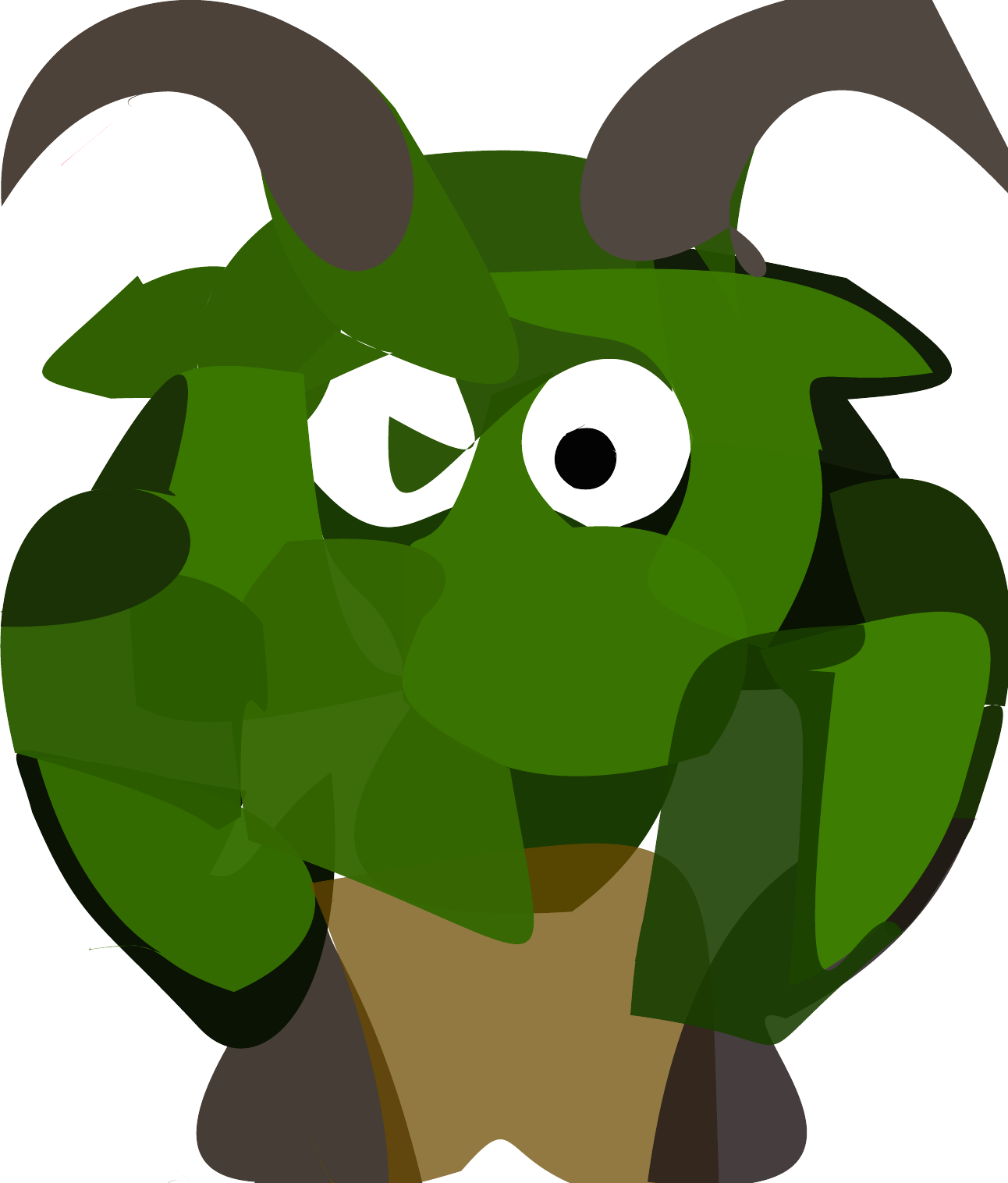}
&
\includegraphics[width=1.65cm]{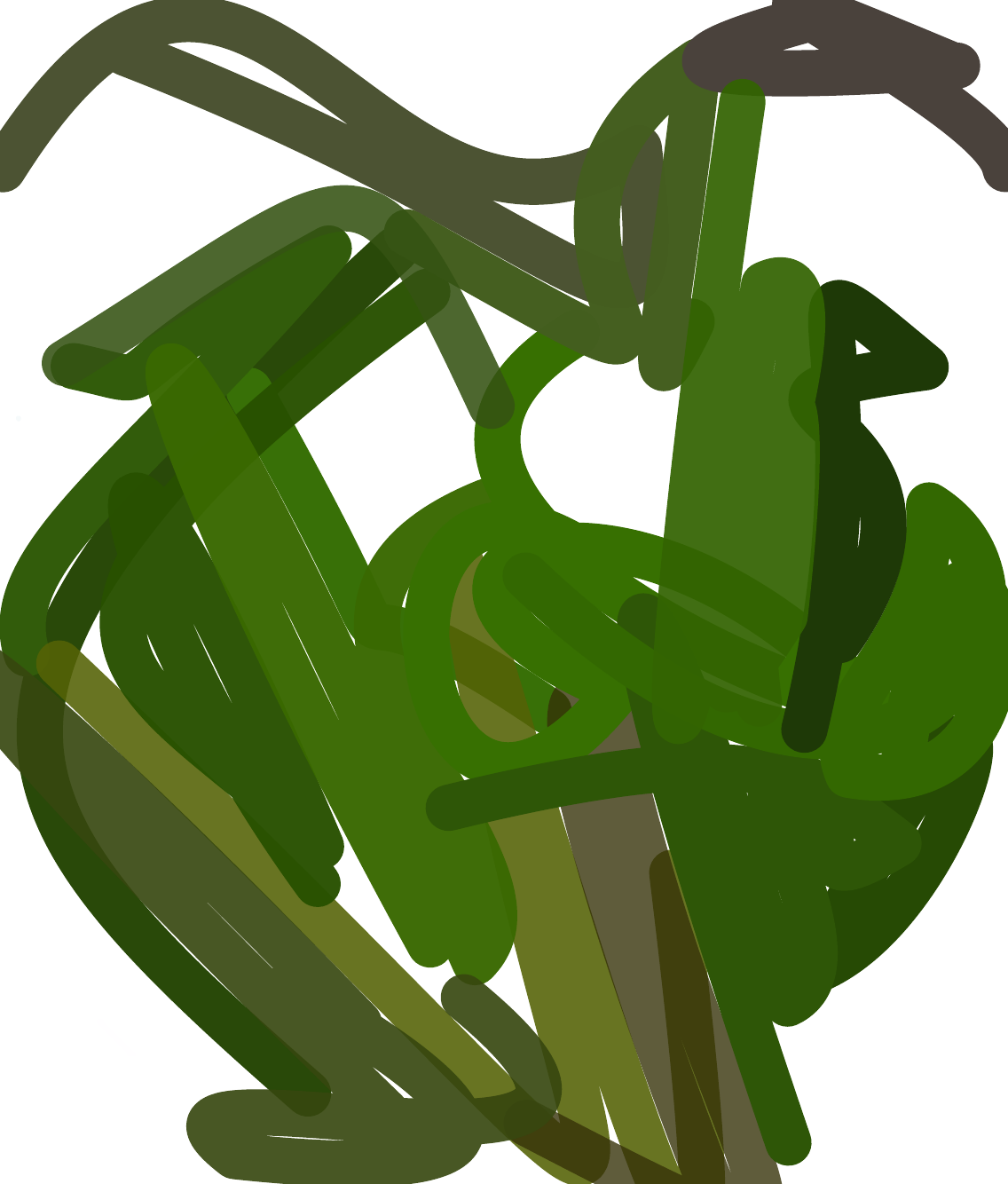}
&
\includegraphics[width=1.65cm]{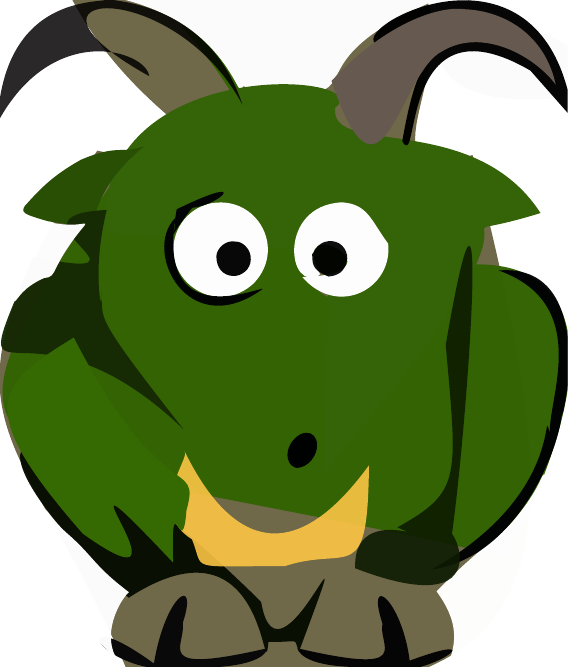}
\\
% 2
Medium Vector
&
\includegraphics[width=1.65cm]{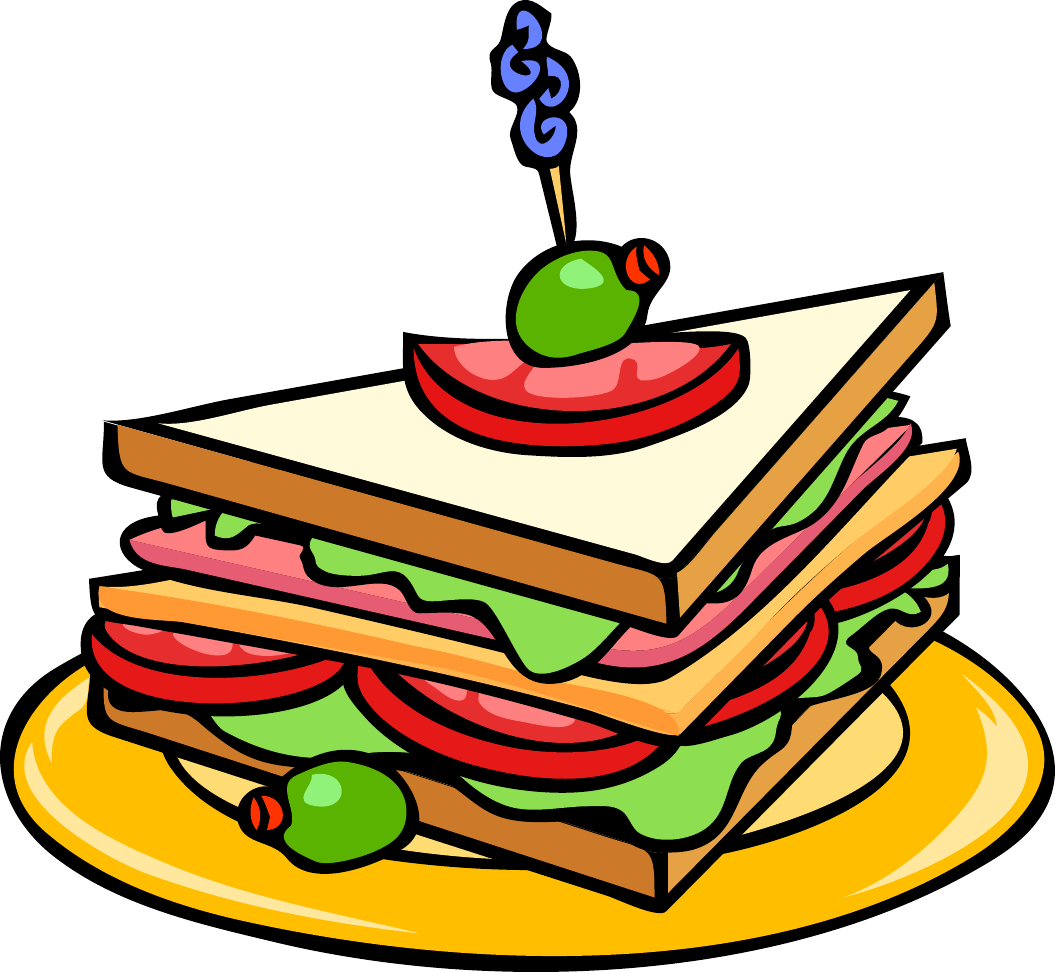}
&
\includegraphics[width=1.65cm]{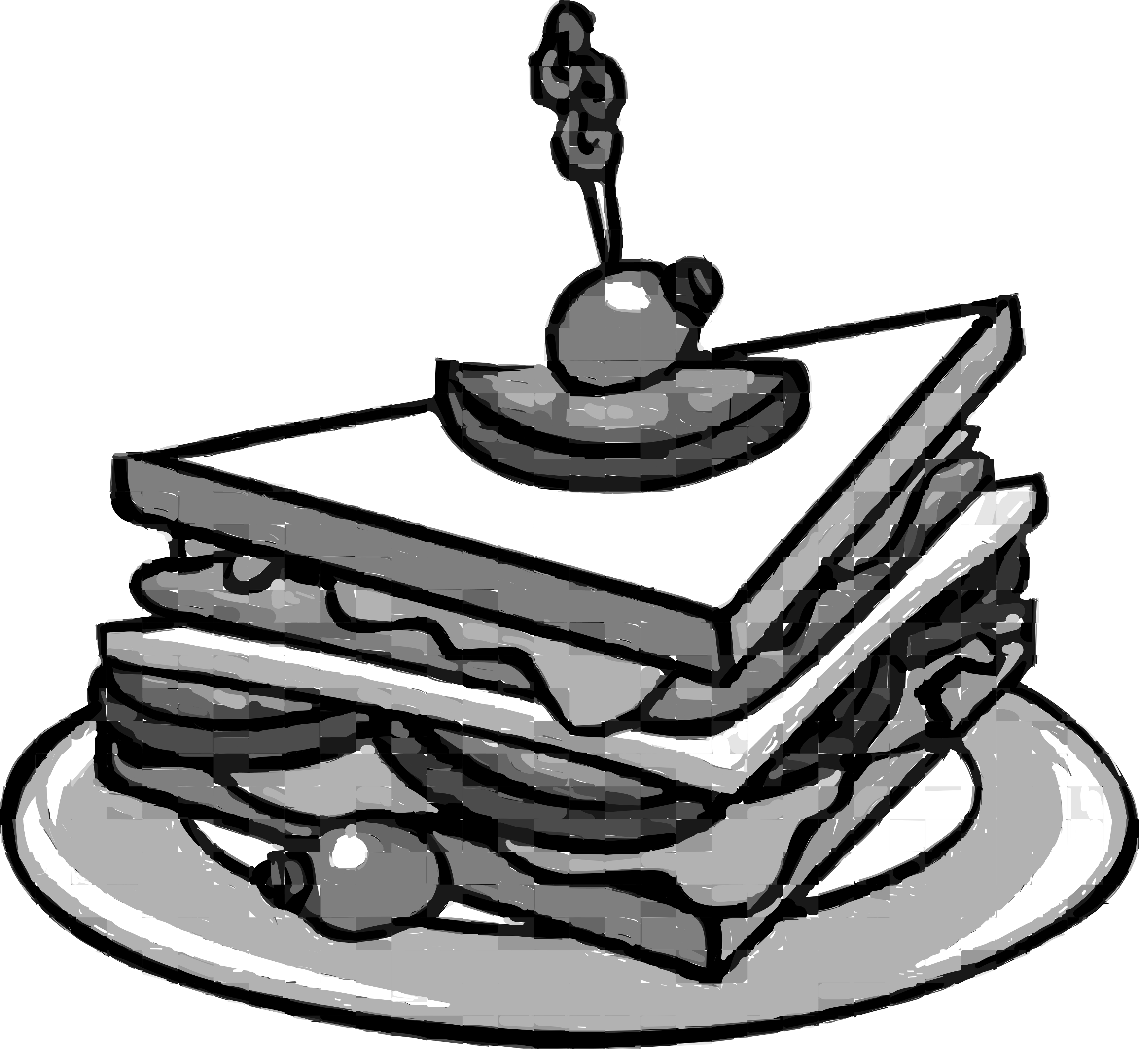}
&
\includegraphics[width=1.65cm]{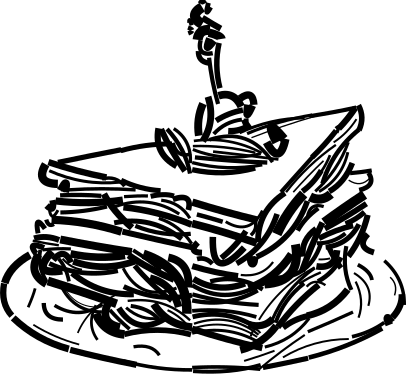}
&
\includegraphics[width=1.65cm]{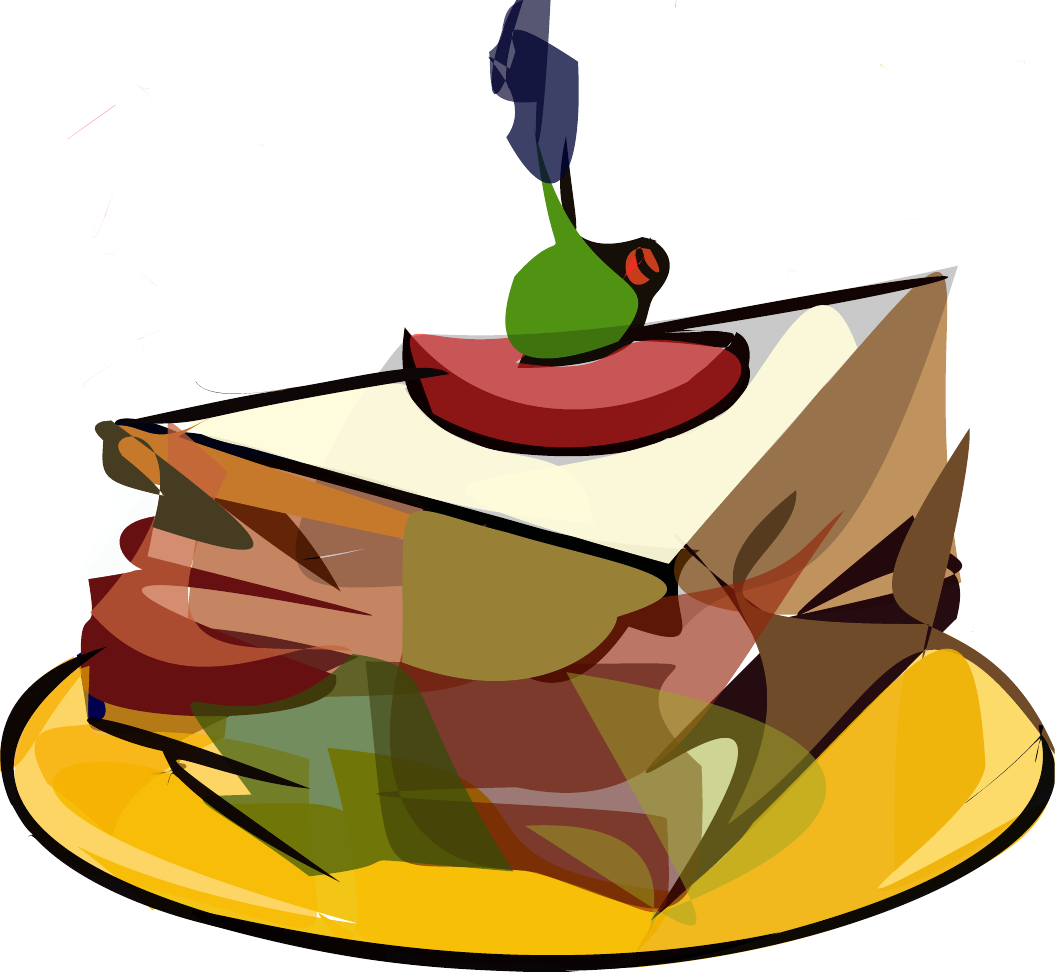}
&
\includegraphics[width=1.65cm]{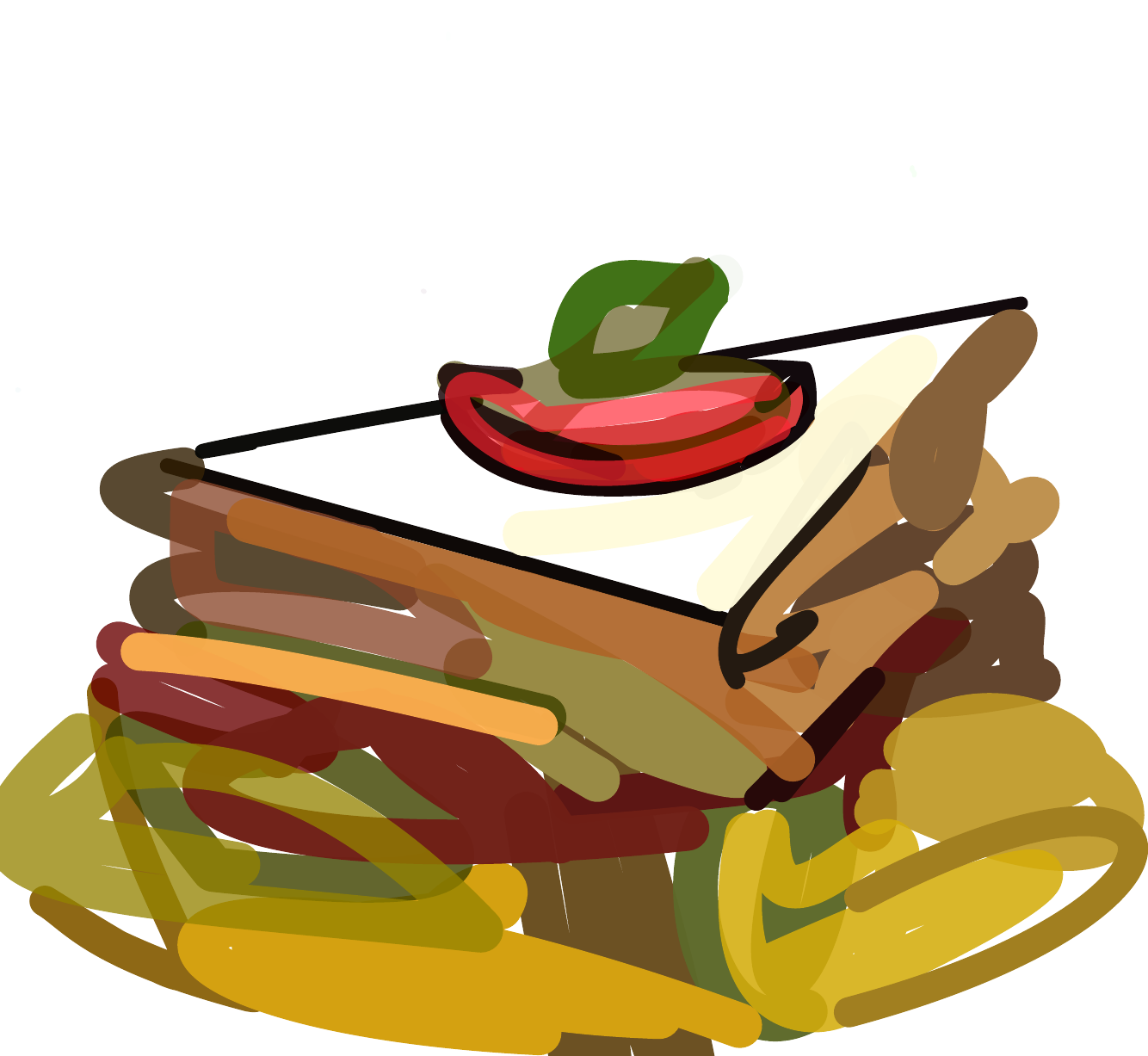}
&
\includegraphics[width=1.65cm]{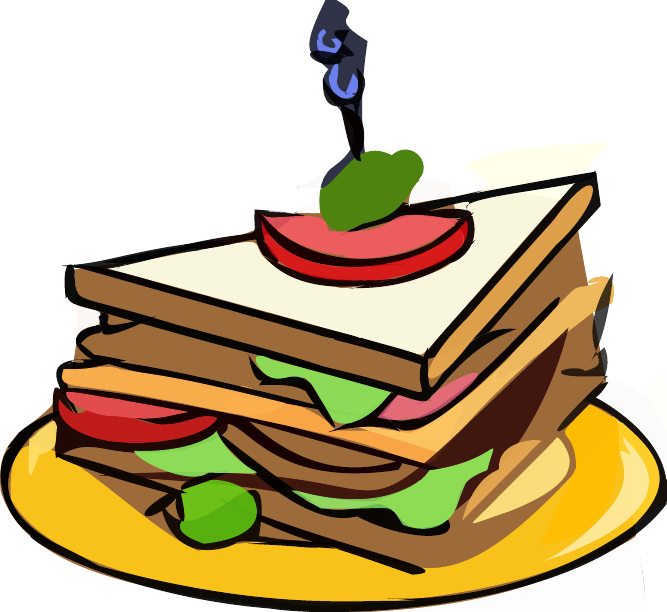}
\\
% 3
Hard Vector
&
\includegraphics[width=1.65cm, height=1cm]{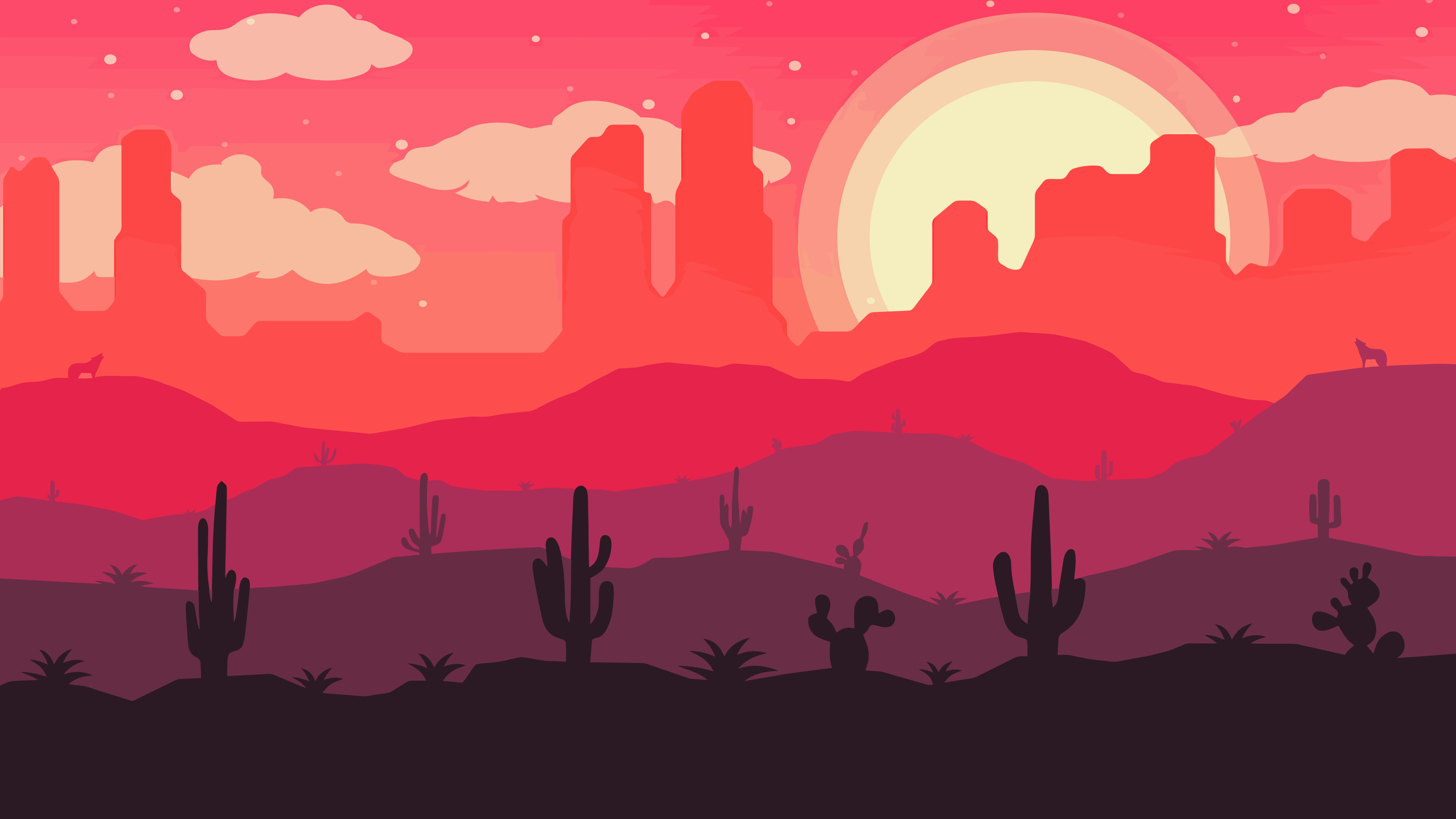}
&
\includegraphics[width=1.65cm, height=1cm]{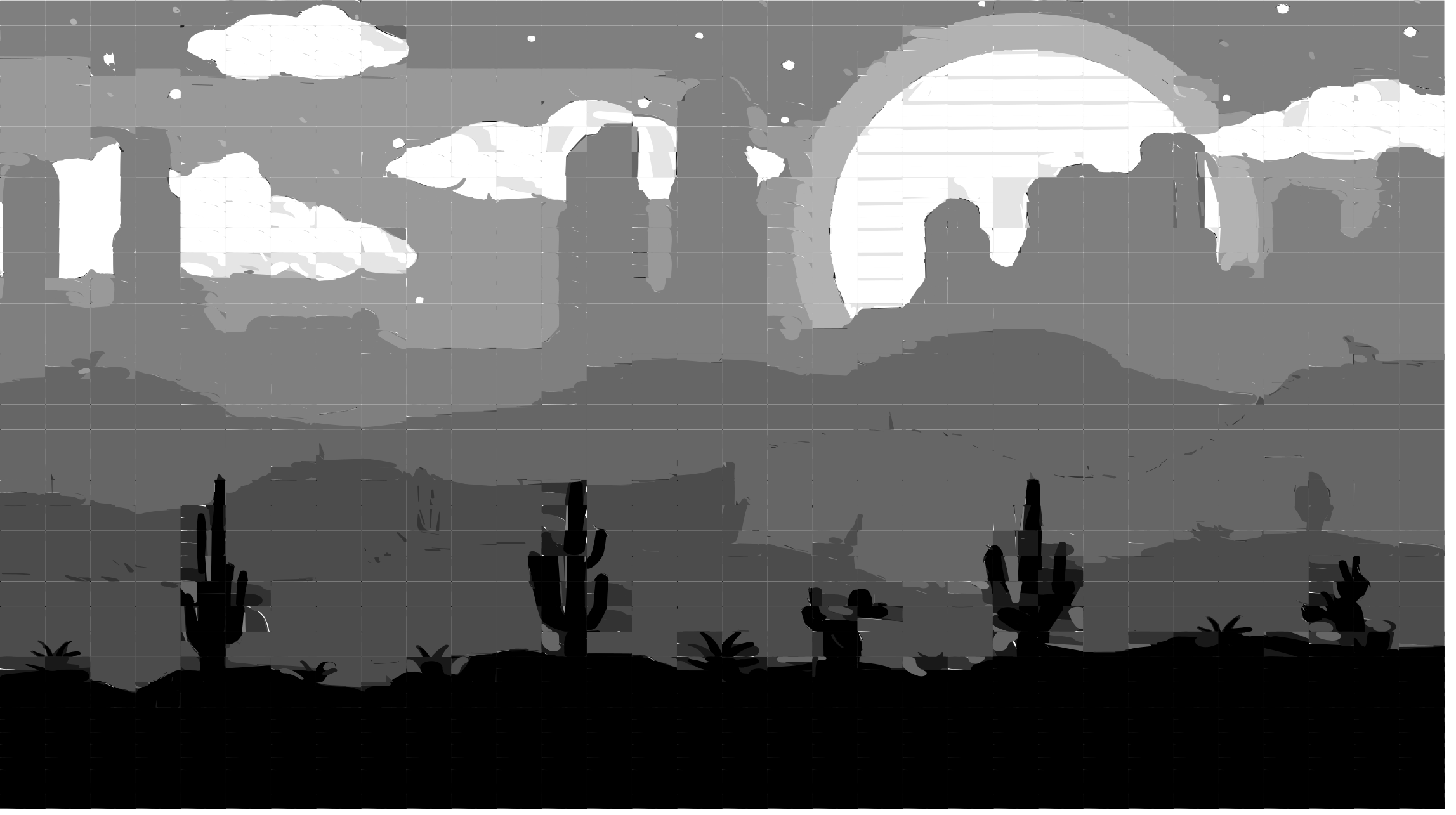}
&
\includegraphics[width=1.65cm, height=1cm]{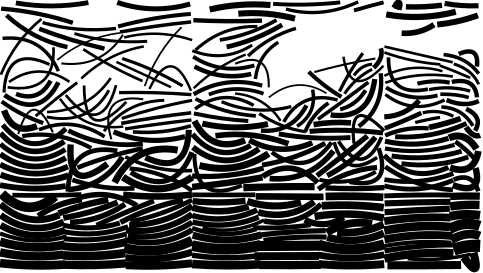}
&
\includegraphics[width=1.65cm, height=1cm]{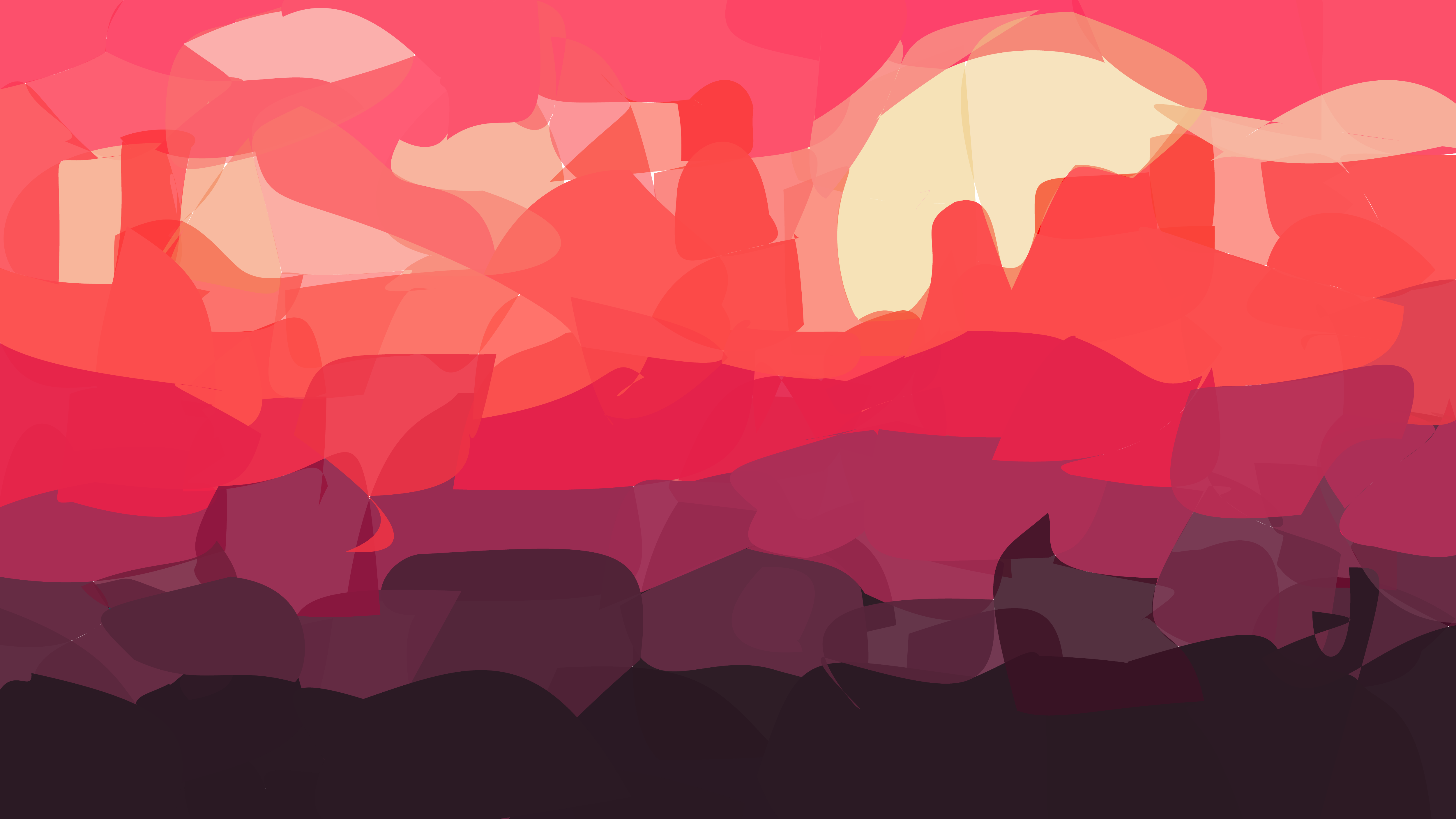}
&
\includegraphics[width=1.65cm, height=1cm]{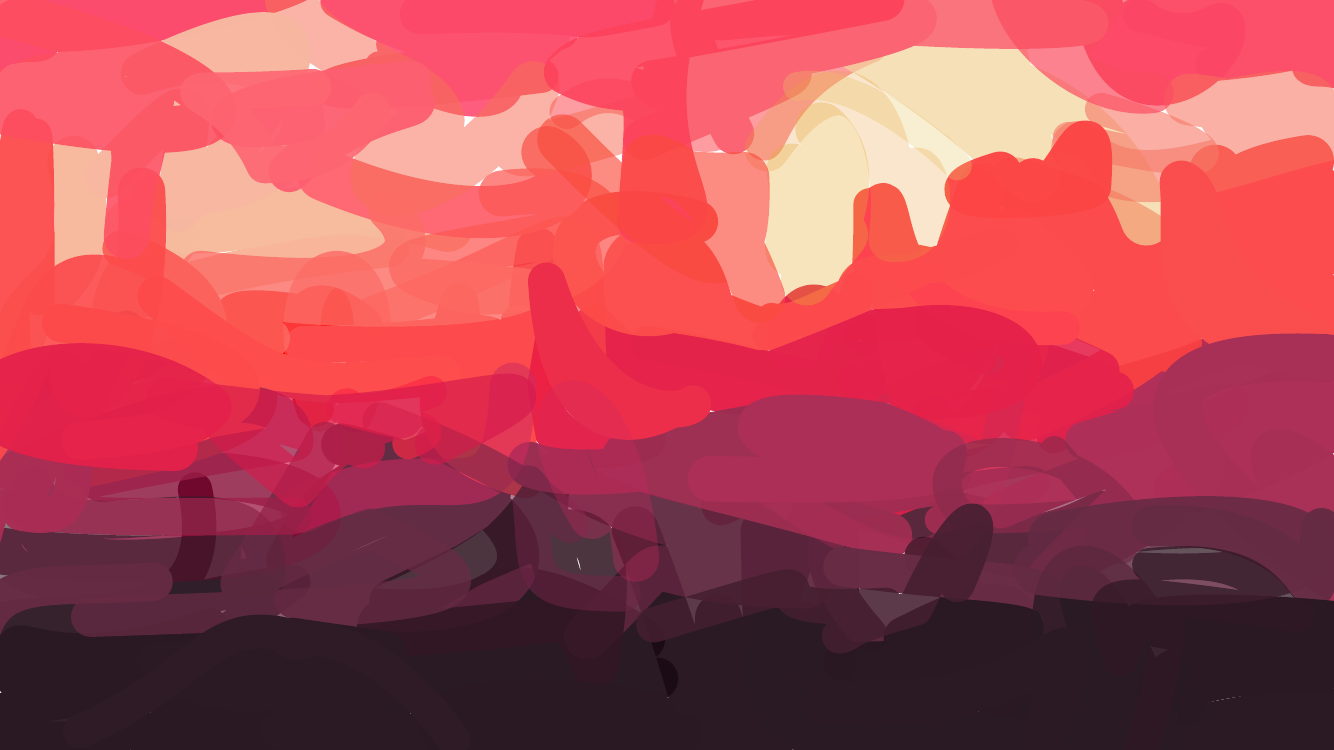}
&
\includegraphics[width=1.65cm, height=1cm]{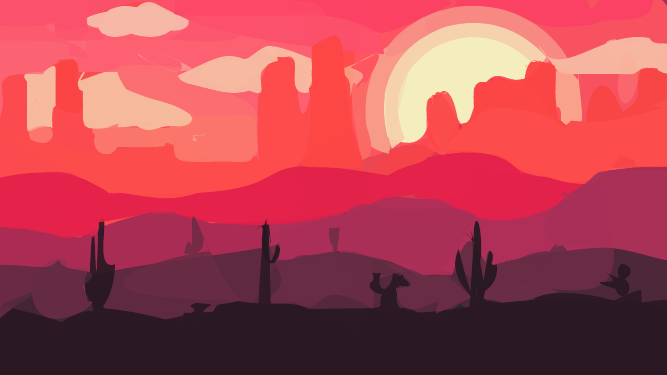}
\\
% 4
Simple Raster
&
\includegraphics[width=1.65cm]{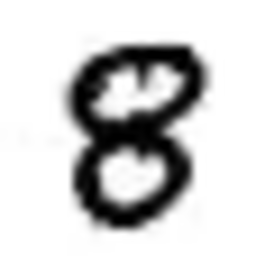}
&
\includegraphics[width=1.65cm]{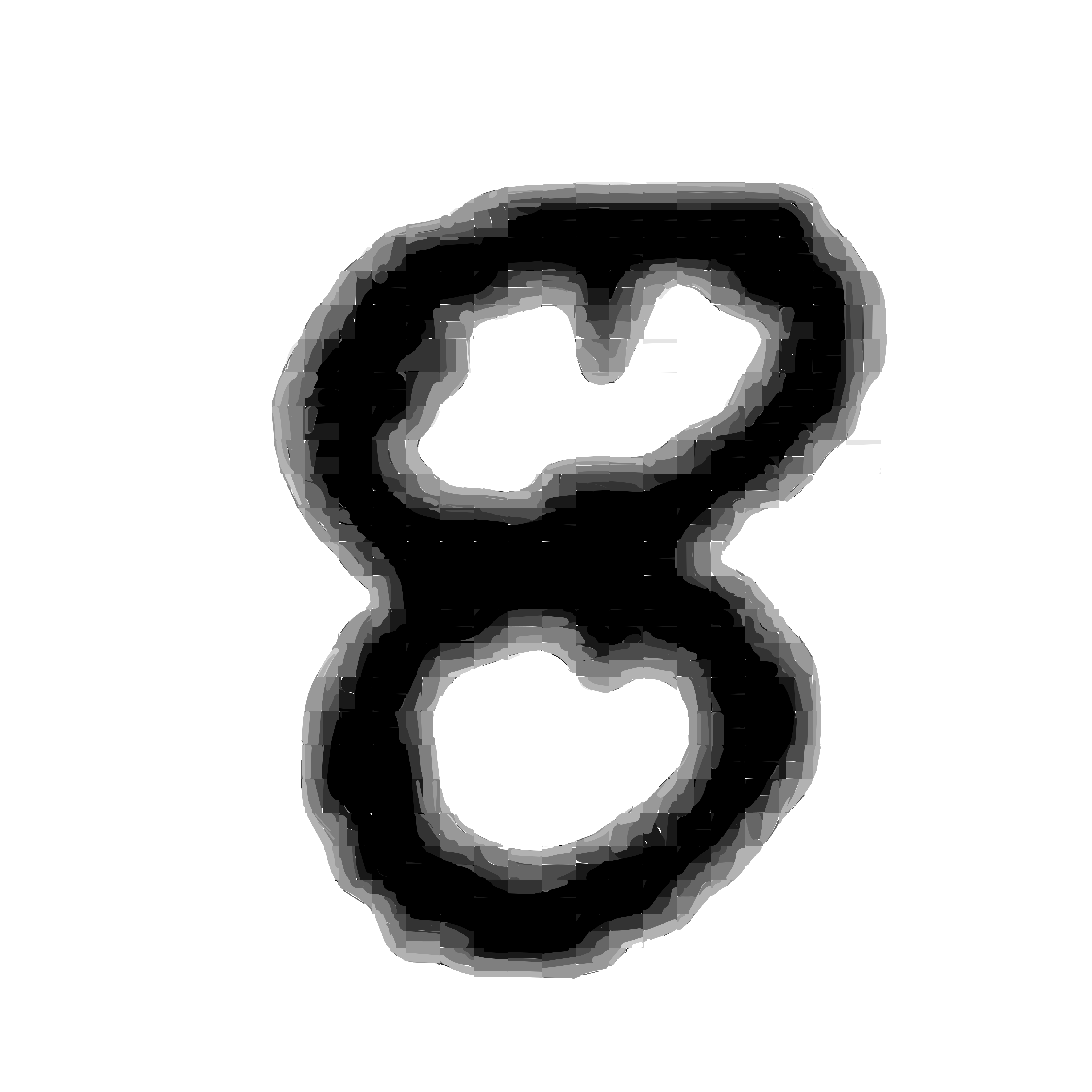}
&
\includegraphics[width=0.9cm, height=1.25cm]{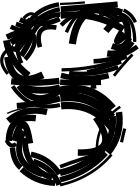}
&
\includegraphics[width=1.65cm]{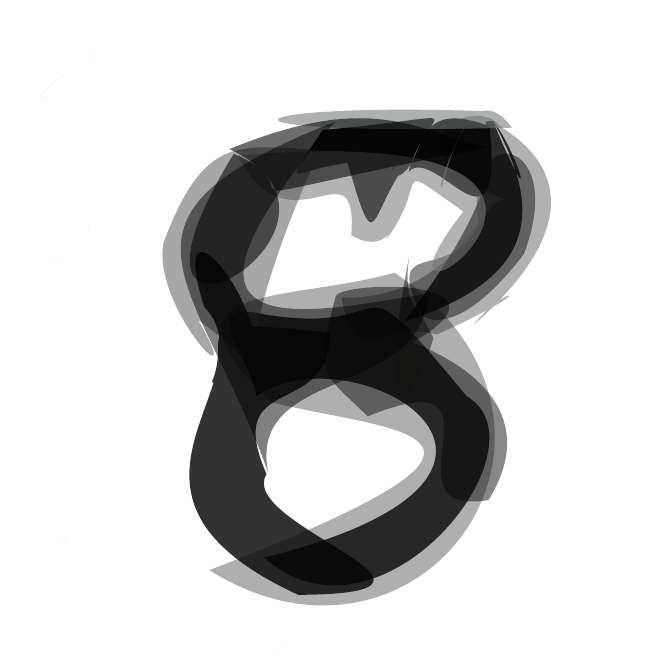}
&
\includegraphics[width=1.65cm]{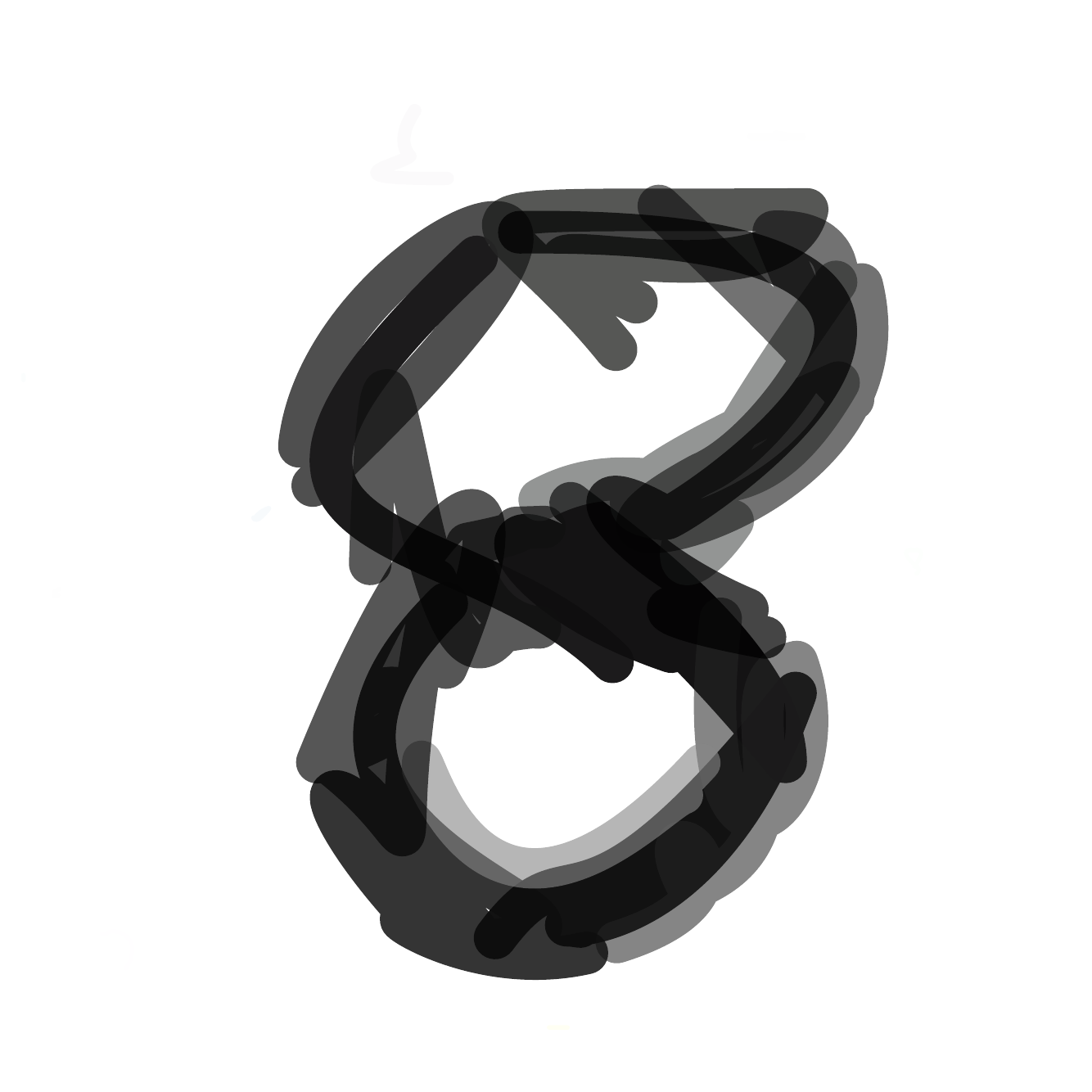}
&
\includegraphics[width=1.65cm]{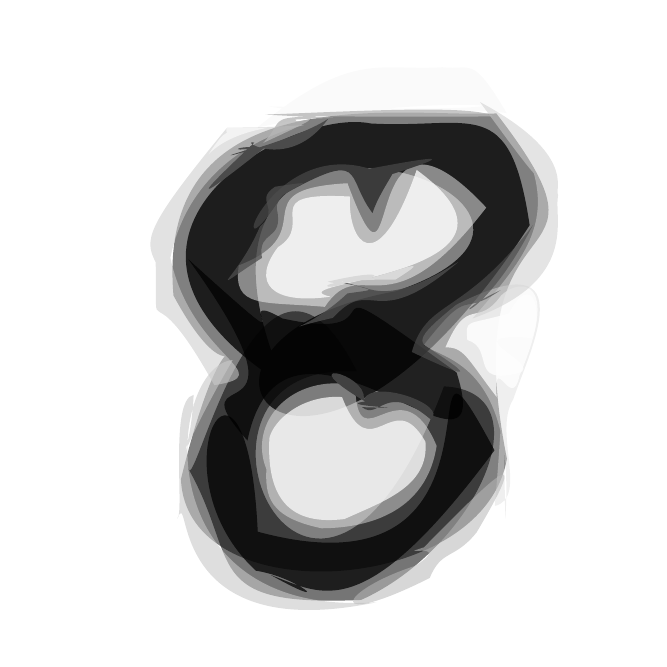}
\\
% 4
Medium Raster
&
\includegraphics[width=1.65cm]{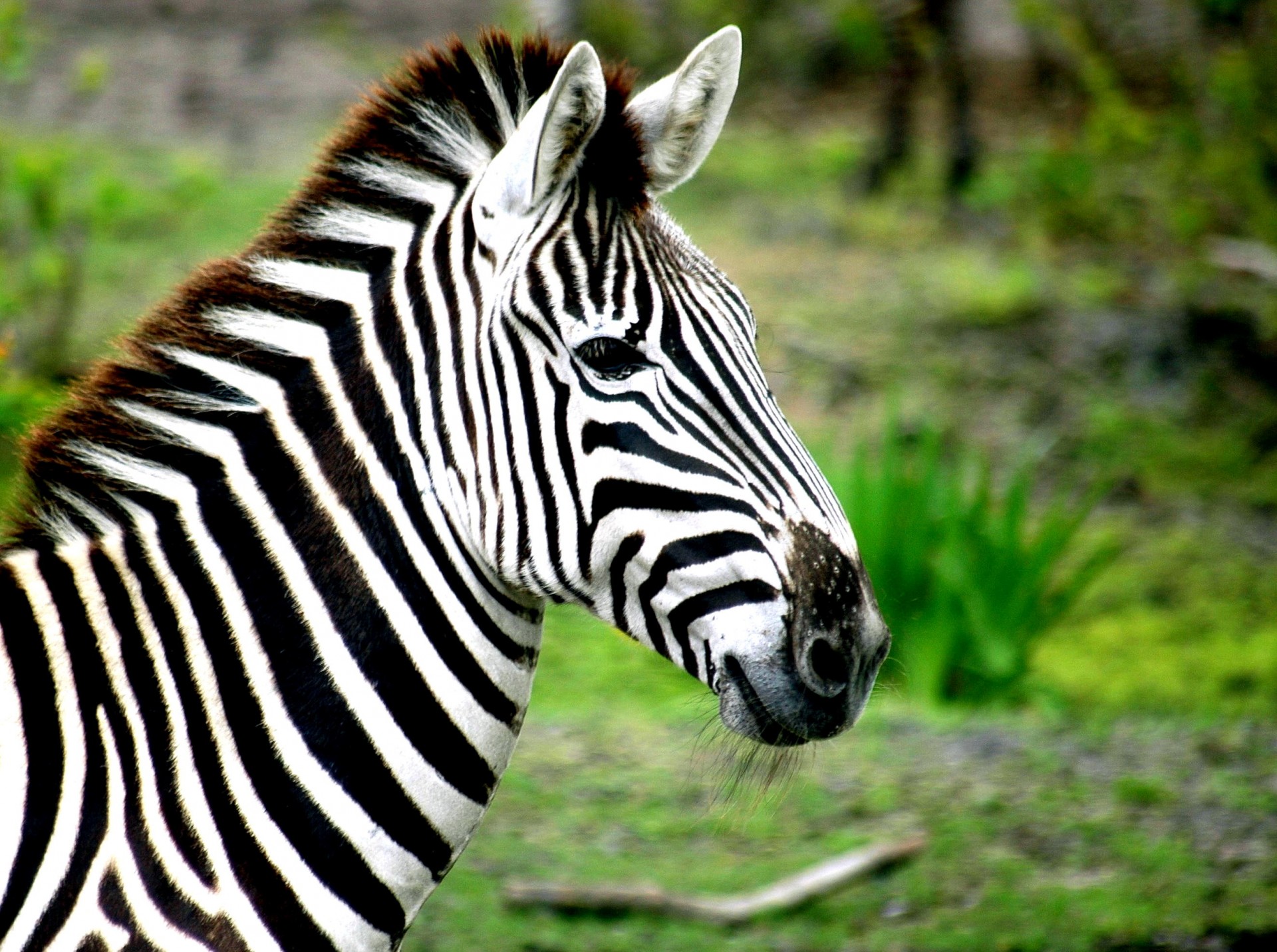}
&
\includegraphics[width=1.65cm]{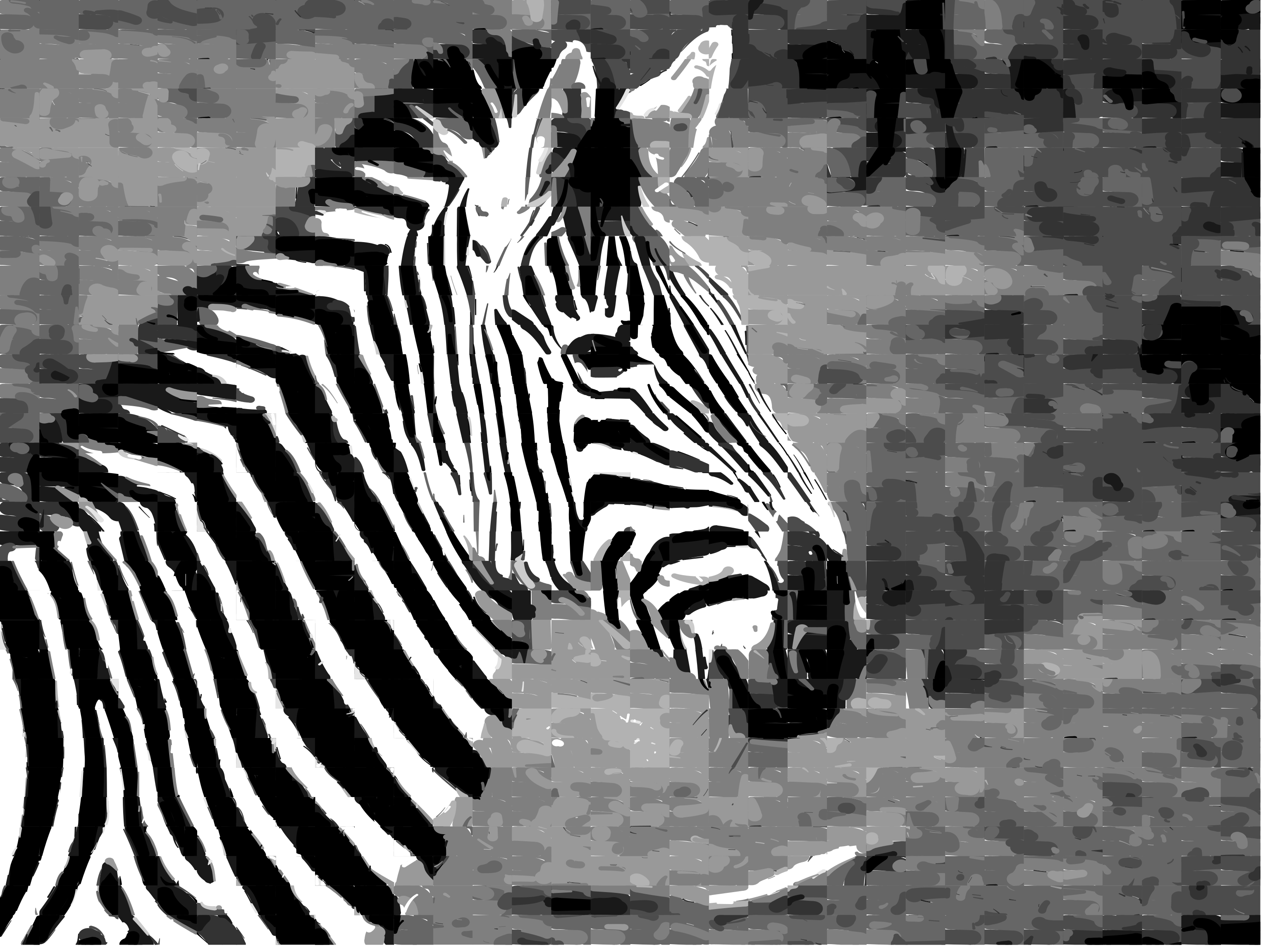}
&
\includegraphics[width=1.65cm]{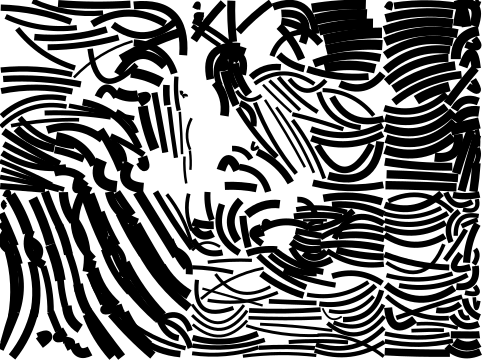}
&
\includegraphics[width=1.65cm]{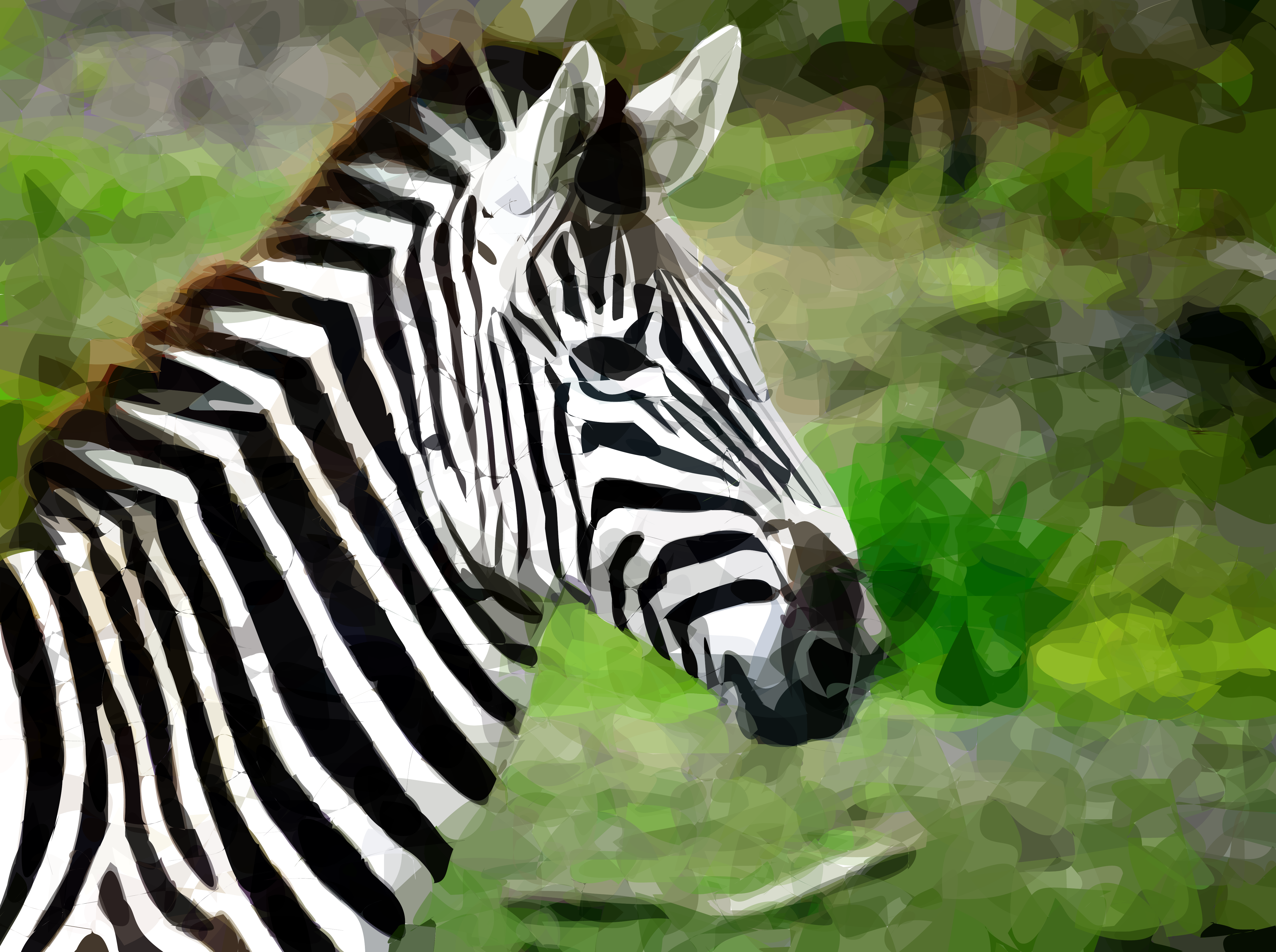}
&
\includegraphics[width=1.65cm]{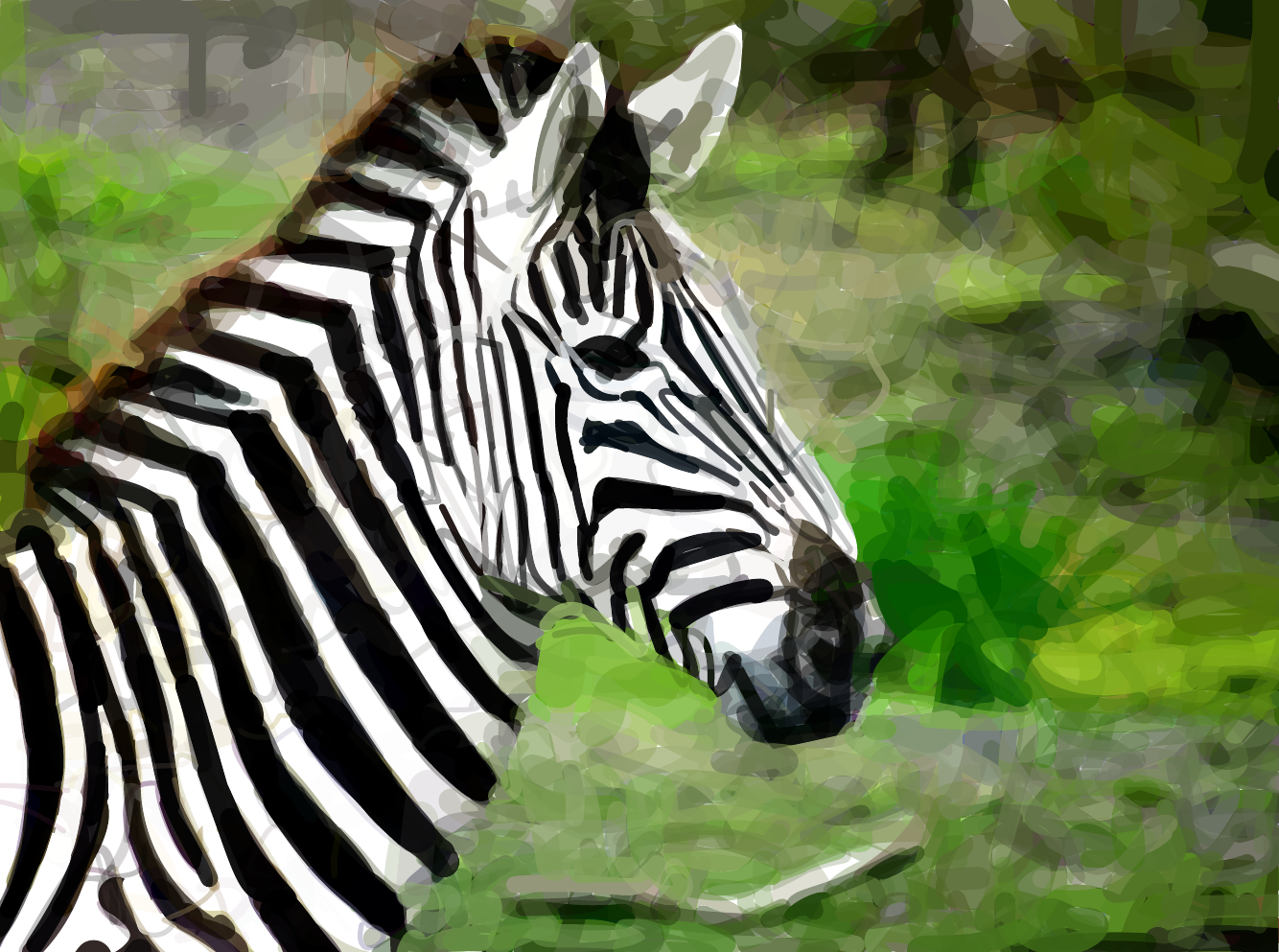}
&
\includegraphics[width=1.65cm]{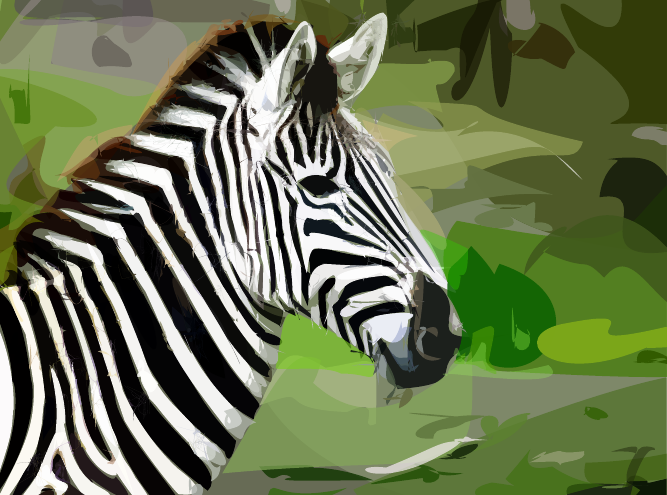}
\\
% 3
Hard Raster
&
\includegraphics[width=1.65cm, height=1.2cm]{images/original/landscape_medium.png}
&
\includegraphics[width=1.65cm, height=1.2cm]{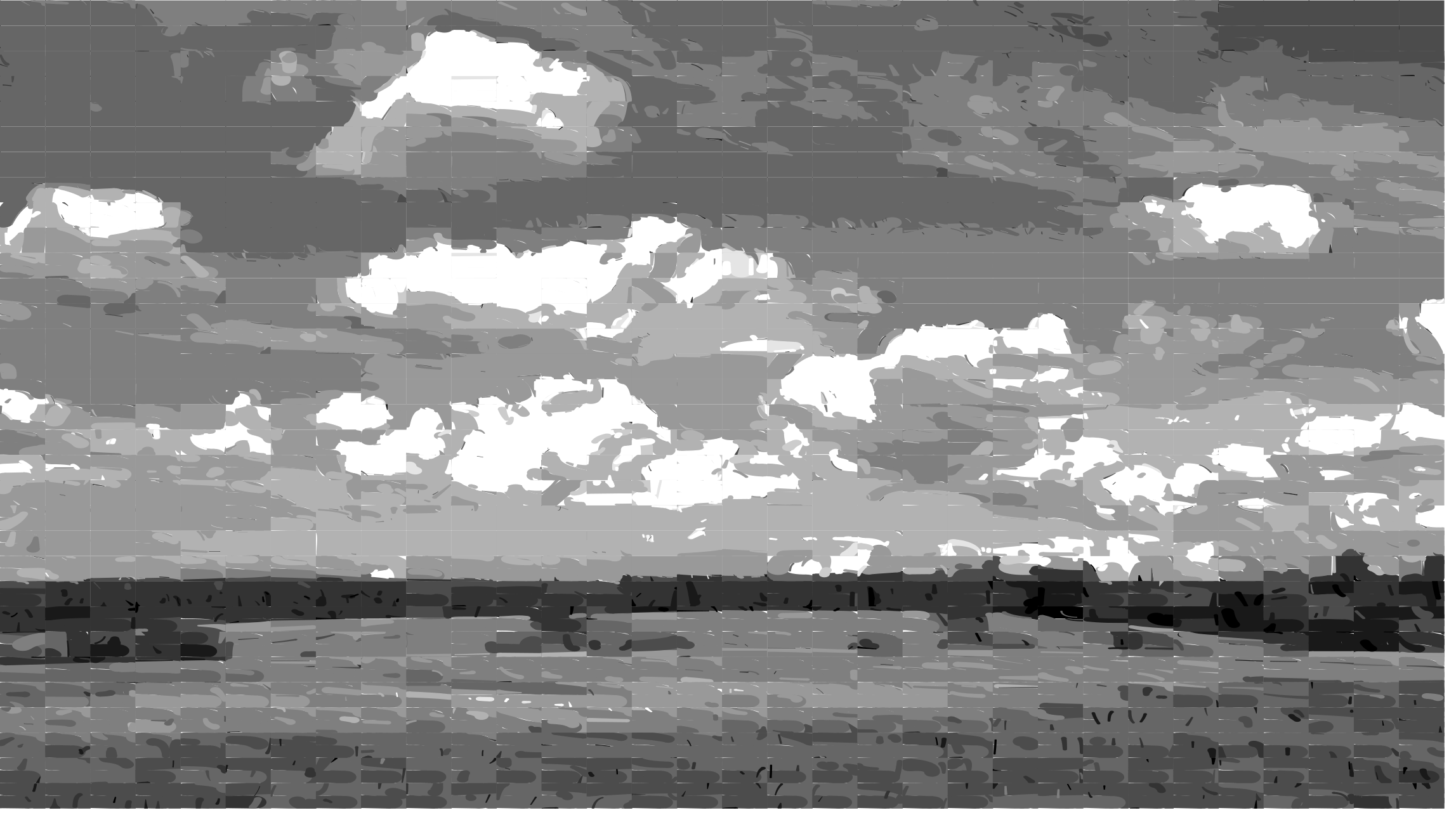}
&
\includegraphics[width=1.65cm, height=1.2cm]{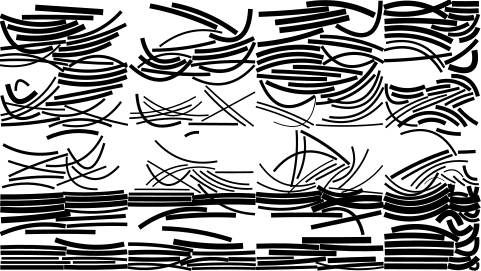}
&
\includegraphics[width=1.65cm, height=1.2cm]{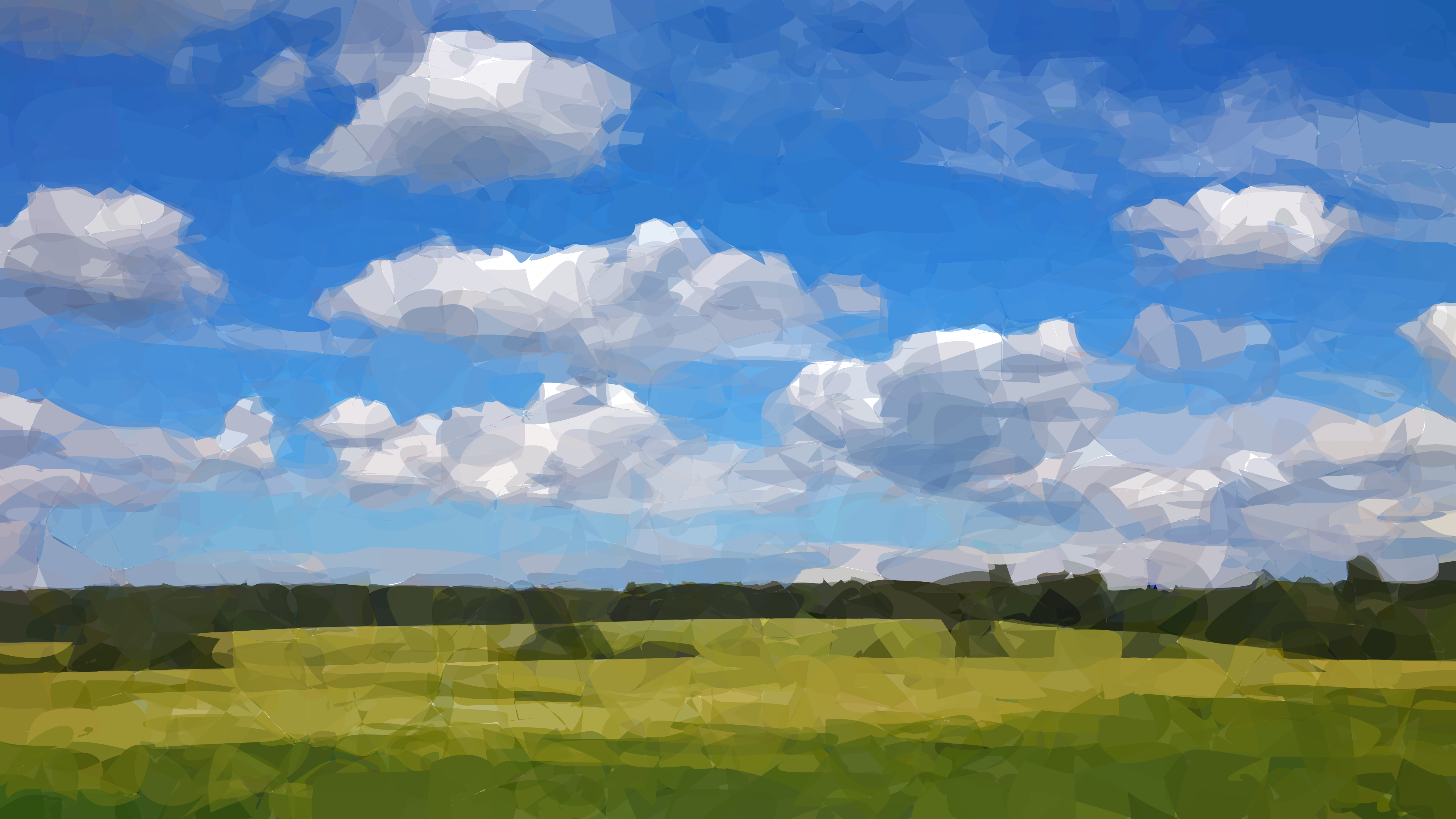}
&
\includegraphics[width=1.65cm, height=1.2cm]{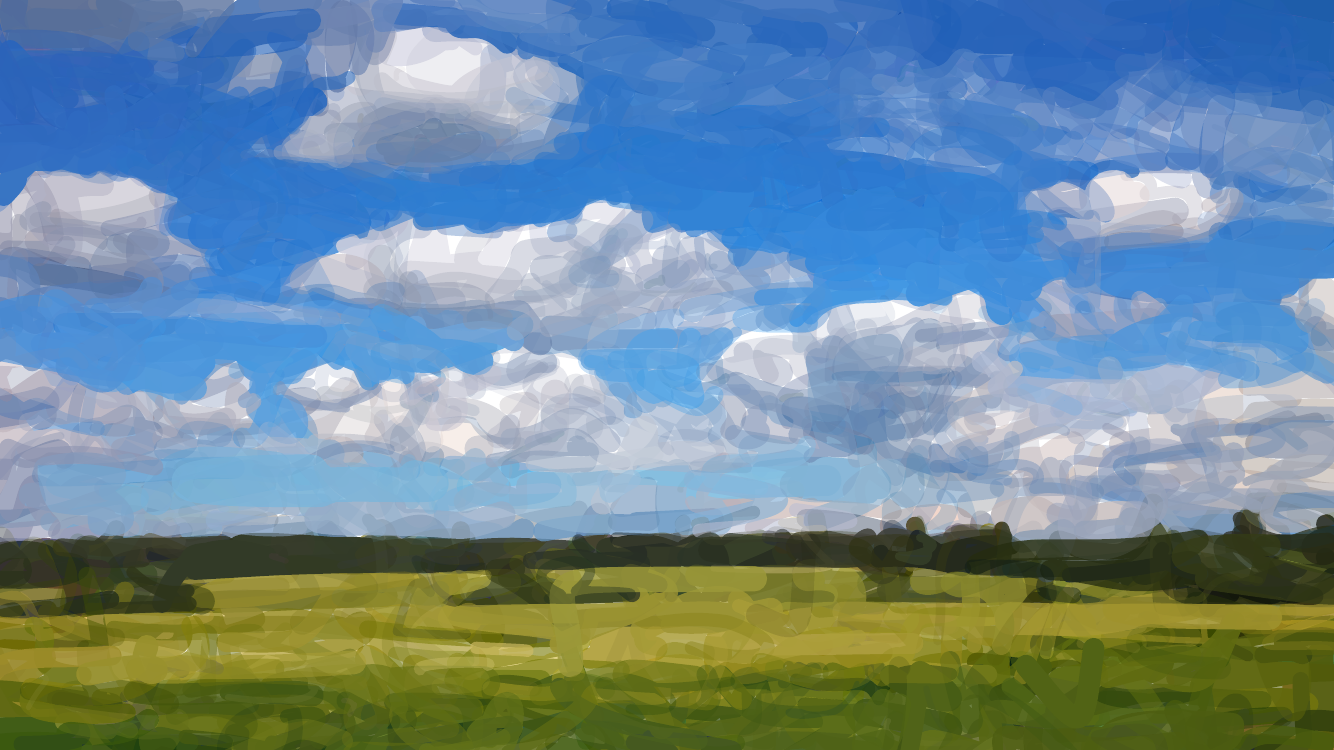}
&
\includegraphics[width=1.65cm, height=1.2cm]{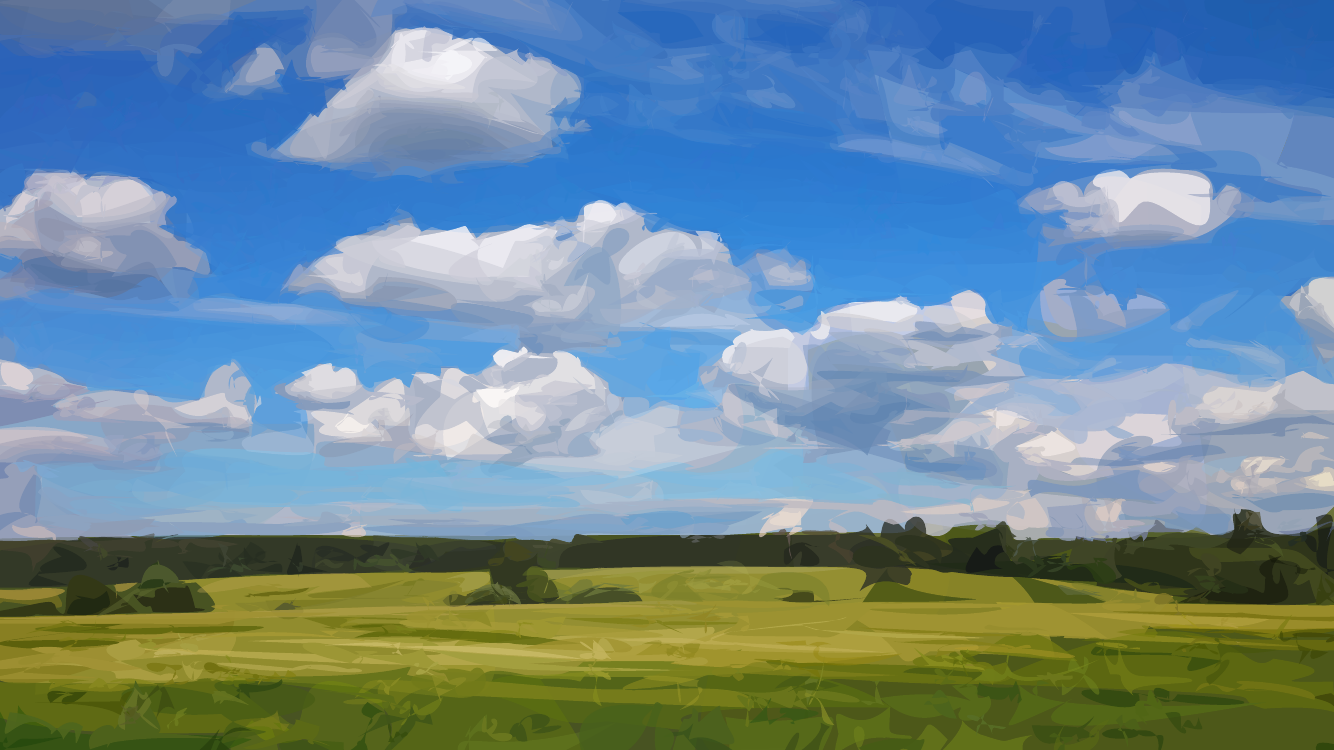}
\\
\bottomrule
\end{tabular}
\end{center}
\caption{Qualitative comparisons of image vectorization results using different methods. DiffVG closed stands for the DiffVG method with closed paths, unclosed - with unclosed strokes. DiffVG and LIVE results for the $3$ initially vector images have the paths amount as in the original images, for the Simple raster image $32$ paths were used, for the Medium and Hard Raster -- $1024$ paths. The other methods have been run with their default parameters.}
\label{fig:results-comparing}
\end{figure}

% Какие критерии еще раз и почему именно они
\subsection{Comparison Criteria}

To make a valuable comparison of vectorization methods, it is necessary to consider that the visual appealingness and similarity of the resulting vector image to the original raster image is not the only important criterion. 
The speed of vectorization is also an important factor.

The main advantages of a vector image are its simplicity and a small number of shapes used. 
Although a vector image can contain various shapes (circles, rectangles, paths, etc.), vectorization methods tend to generate images using only paths. 
Paths themselves consist of segments (B\`ezier curves, straight lines, etc.) and their number in each path should be low as well.
This is necessary both for simpler post-processing by designers and faster image transfer to the user through the Internet with subsequent vector image rendering.

Thus, the main five criteria for evaluating vectorization methods are: 
\begin{enumerate}
    \item similarity to the original bitmap;
    \item the simplicity or complexity of the resulting image including the number of shapes and their parameters;
    \item the speed of generation;
    \item versatility~--- the ability to generate a fairly accurate copy of the input image without prior model training; 
    \item human control to adjust hyperparameters.
\end{enumerate}

However, taking into account all the criteria at the same time is challenging, because the methods we consider have many different parameters that affect all the criteria simultaneously. 
Typically, by changing one parameter, one can achieve an increase in image processing speed but at the same time reduce the quality of the resulting image.

\subsection{Experiment Setup}

%In this section, we provide details of our experiments and describe the approaches and parameter values that seemed to be the most optimal when applying the methods.

We selected $6$ images for comparison: $3$ rasterized vector images and $3$ bitmaps of different complexity. 
The original target vector images before rasterization had the following number of paths: dragon had $25$, burger -- $62$, red landscape -- $100$. 
Ideally, vectorization methods should create images consisting of an approximately similar number of paths in a short period of time. 
%Ideally, on rasterized vector images, vectorization methods should produce an exact vector copy with approximately the original number of shapes in a small amount of time. 
At the same time, it does not worth expecting vectorization methods to account for every image detail on initially raster images, as an over-detailed vector image does not satisfy the simplicity criterion~--- a smaller number of paths. 
It is also desirable that simple monochrome patches should be decorated with a minimum number of shapes and the image subject should not be lost.

%Rasterized vector images are easier to reconstruct. 
We compare the following methods (with publicly available implementation): Mang2Vec, Deep Vectorization of Technical Drawings (DVoTD), iterative DiffVG and LIVE methods, and online methods. 
The following models are not included in the comparison: 
1) Im2Vec~\cite{reddy2021im2vec}, because we could not confirm in practice the results described in the paper, and it also requires additional serious pre-training for processing relatively diverse images; 
2) ClipGEN~\cite{shen2021clipgen}, because the model is limited to the set of predefined classes and there no its implementation is publicly available; 
3) VAE and GAN introduced in DiffVG~\cite{li2020diffvg}, because they also require additional pre-training. Also, even on such a simple dataset as MNIST, we found their results are not satisfactory enough; 
4) algorithmic methods, since we found no implementations publicly available.
Fig.~\ref{fig:results-comparing} contains the original images and their vectorized versions using different methods reviewed in our paper. 

Our experiments have proven that the Mang2Vec and DVoTD models are not versatile, since they are capable of processing only black-and-white images.
At first glance, Mang2Vec vectorizes the image well, but its significant drawback is the use of a very large number of shapes: for instance, the ``burger'' image on the last 5th iteration had $3065$ paths and the ``dragon'' image on the 20th iteration $16600$ had paths. 
Also, the method adds many <clippath> and <circle> tags, which seems useless.
The method uses image splitting into patches and performs a separate vectorization of each patch, which is acceptable when processing detailed manga images. 
However, a large monochrome space becomes divided into a large number of shapes, which is unacceptable. 
In the Mang2Vec method, you can specify a different number of iterations, but with a small number of them, the patches boundaries, into which the division is performed, become clearly visible.
Since Mang2Vec automatically resizes image to 4096x4096 resolution its working time is constant and is $157$ seconds.

Deep Vectorization of Technical Drawings (DVoTD) was meant to be able to use either quadratic B\'ezier curves or straight lines. 
We managed to run the method for curves, but the implementation of the second method (straight lines) is imperfect and the code is likely to contain some issues that lead to a crash during execution, which we could not fix. 
The method struggles to fill in contours with a solid color, as it was originally made to generate black stroke lines. 
We ran the method with default parameters and, for instance, the ``dragon'' image had $132$ paths.
The running time of the 270x480 image was $142$ seconds, 373x405 -- $179$ seconds, 582x496 -- $273$ seconds.

Vectorization by the DiffVG iterative method can be done in two ways: generating images consisting of curves and of closed shapes. 
The approach is simple and quite effective, but it produces many artifacts with shapes that the method apparently attempts to hide, but it fails to succeed. 
In addition, the reconstruction of absolutely exact visual copies of the original vector images cannot be obtained even using a large number of paths ($1024$ shapes).

% DiffVG closed path running time on GPU for 500 iterations:
% Dragon (496x582) 25 paths = 56 sec
% Burger (405x373) 62 paths = 48 sec
% Scene1 (1920x1080) 32 paths = 4 min 56 sec
% Scene1 (1920x1080) 64 paths = 5 min 19 sec
% Scene1 (1920x1080) 100 paths = 5 min 49 sec
% Scene1 (1920x1080) 256 paths = 7 min 15 sec
% Scene1 (1920x1080) 512 paths = 9 min 43 sec
% Scene1 (1920x1080) 1024 paths = 13 min 57 sec
% Zebra (1920x1432) 256 paths = 9 min 42 sec
% Zebra (1920x1432) 512 paths = 12 min 50 sec
% Zebra (1920x1432) 1024 paths = 18 min 19 sec
% Landscape_medium (1920x1080) 256 paths = 7:18
% Landscape_medium (1920x1080) 512 paths = 9:35
% Landscape_medium (1920x1080) 1024 paths = 13:54

% Тут еще будет время работы для DiffVG unclosed path:
% ...

% вычитать текст ниже...
Different renderers convert a vector image to a bitmap in different ways. 
For example, images generated by DiffVG and LIVE will look inaccurate and careless, when they are rendered by the InkScape. This behavior occurs due to the fact that these images contain curves protruding beyond the edges of the viewBox attribute, and InkScape displays them instead of cropping them.
% Images generated by DiffVG and LIVE look inaccurate and careless in Fig.~\ref{fig:results-comparing}, when they are rendered by the InkScape, which was used to generate the PDF of the paper you are reading. 
At the same time, other renderers, for example, in Google Chrome browser, process images correctly without extra curves that remain beyond the viewBox. 
However, these curves are still a problem, since they are superfluous and they create additional artifacts in the image code and add an extra size to it.

% дальше не вычитали
The LIVE method iteratively adds a layer consisting of one or more shapes specified by user to the image and optimizes the resulting image. 
In addition, number of image processing iterations after applying each layer should be specified manually. 
LIVE sets the number of iterations to $500$ by default, but we noticed that after about $200$ iterations, the image almost does not change, so we set this value in our experiments. 
In the optimization process, LIVE uses the DiffVG rasterizer to convert the current vector image into a raster image and compare it with the target image.
Rasterization is performed by default at the same resolution as the target image, but for large resolutions it is computationally time-consuming. 
For example, for a resolution of 1080x1920, processing the first layer in $200$ iterations on the Nvidia RTX 3090 Ti GPU took $212$ seconds, then by 10th layer, processing of one layer reached $244$ seconds. Finally, processing of $28$ layers has took almost $2$ hours. Therefore, we decided to pre-scale the raster images so that their maximum side does not exceed $512$ pixels. 
At the same time, it is worth noting that this approach carries the risk of losing details in the image. With this approach, the processing of the first layer took $16$ seconds, but by the 10th layer, the processing time of one layer was $40$ seconds. 
The total processing time of $46$ layers took about $32$ minutes. 
When limiting the maximum image side to $256$ pixels, the processing time of the first layer was $6$ seconds, and totally $32$ layers were processed in $22$ minutes. 
It should also be known that the processing time is also affected by the number of shapes in each layer. In our experiments we used less than $7$ shapes in each layer only the first $5$ layers, after that we used $7$, $10$, $20$ or $30$ shapes in each layer gradually increasing the number.

LIVE creates the most accurate images among the ML methods. However, the most significant disadvantage of this method is a very long image processing time, which depends on the number of layer additions, the number of iterations of processing each layer, the dimensions of the image for which intermediate rasterization is performed.

It is worth noting that DiffVG also has problems with long intermediate rasterization, however, due to the smaller total number of iterations, this is less noticeable. 
The results of DiffVG and LIVE operation times are shown in more detail in the Tab.~\ref{tab:diffvg-times} and Tab.~\ref{tab:live-times}.

\begin{table}
\begin{center}
\begin{tabular}{| c | c | c |}
\hline
Image Resolution & Total paths & Time (sec)\\
\hline
512x512 & 16 & 47 \\
405x373 & 62 & 48 \\
496x582 & 25 & 56 \\
512x512 & 32 & 56 \\
512x288 & 256 & 143 \\
512x381 & 256 & 186 \\
512x288 & 512 & 216 \\
512x381 & 512 & 280 \\
1920x1080 & 32 & 296 \\
1920x1080 & 64 & 319 \\
1920x1080 & 100 & 349 \\
512x288 & 1024 & 389 \\
1920x1080 & 256 & 435 \\
512x381 & 1024 & 446 \\
1920x1432 & 256 & 582 \\
1920x1080 & 512 & 583 \\
1920x1432 & 512 & 770 \\
1920x1080 & 1024 & 837 \\
1920x1432 & 1024 & 1099 \\
\hline
\end{tabular}
\end{center}
\caption{The running time of the DiffVG iterative algorithm at different startup parameters on NVidia RTX 3090Ti GPU. Total iterations number is $500$. There is no serious speed difference between methods with closed and unclosed paths.}
\label{tab:diffvg-times}
\end{table}

% LIVE running time on GPU for 200 iterations:
% (swaped resolution)
% Scene1 (144x256) 100x1 paths 100 layers = 20 min 31 sec
% Burger (235x256) 62x1 paths 62 layers = 15 min 45 sec
% Dragon (256x218) 25x1 paths 25 layers = 4 min 38 sec
% Scene1 (144x256) 1x1+3x1+5x1+7x8=64 paths 11 layers= 2 min 01 sec
% Burger (235x256) 1x1+3x1+5x1+7x8=64 paths 11 layers = 2 min 47 sec
% Dragon (256x218) 1x1+3x1+5x1+7x8=64 paths 11 layers = 2 min 40 sec
% Scene1 (144x256) 1x1+3x1+4x1+5x1+7x1+10x8=100 paths 13 layers = 2 min 49 sec
% Burger (235x256) 1x1+3x1+5x2+7x7=62 paths 11 layers= 2 min 42 sec
% Dragon (256x218) 1x1+2x1+3x1+4x1+5x3=25 paths 7 layers = 1 min 18 sec
% Zebra (190x256) 1+3+5+7+8+10*2+20*4+30*30=1024 paths 41 layers = 41 min + 30 sec
% Dragon (256x218) 1+3+5+7+8+10*2+20*4+30*30=1024 paths 41 layers = 38 min + 52 sec

\begin{table}
\begin{center}
\begin{tabular}{| c | c | c | c | c |}
\hline
Image Resolution & Layers schema & Total layers & Total paths & Time (sec)\\
\hline
256x256 & 4x1 & 4 & 4 & 33 \\
256x256 & 8x1 & 8 & 8 & 77 \\
256x218 & 1,2,3,4,5x3 & 7 & 25 & 78 \\
% 144x256 & 1,3,5,7x8 & 11 & 65 & 121 \\
256x256 & 16x1 & 16 & 16 & 158 \\
% 256x218 & 1,3,5,7x8 & 11 & 65 & 160 \\
235x256 & 1,3,5x2,7x7 & 11 & 62 & 162 \\
% 235x256 & 1,3,5,7x8 & 11 & 65 & 167 \\
144x256 & 1,3,4,5,7,10x8 & 13 & 100 & 169 \\
256x218 & 25x1 & 25 & 25 & 278 \\
256x256 & 32x1 & 32 & 32 & 423 \\
235x256 & 62x1 & 62 & 62 & 945 \\
144x256 & 100x1 & 100 & 100 & 1231 \\
471x512 & 1,3,5,7,10x24 & 28 & 256 & 1448 \\
190x256 & 1,3,5,7,8,10x2,20x4,30x30 & 41 & 1024 & 2332 \\
256x218 & 1,3,5,7,8,10x2,20x4,30x30 & 41 & 1024 & 2490 \\
288x512 & 1,3,5,7,8,10x2,20x4,30x30 & 41 & 1024 & 3197 \\
471x512 & 1,3,5,7,8,10x2,20x4,30x30 & 41 & 1024 & 3891 \\
1080x1920 & 1,3,5,7,10x24 & 28 & 256 & 7150 \\
\hline
\end{tabular}
\end{center}
\caption{The running time of the LIVE algorithm with different startup parameters on NVidia RTX 3090Ti GPU. Each layer is processed for $200$ iterations.}
\label{tab:live-times}
\end{table}

The requirement of DiffVG and LIVE models of directly controlling the number and type of applied shapes on the one hand is their advantage, but on the other hand, since there are no good models for determining the required number of shapes, automatic vectorization of a large set of various raster images becomes almost impossible.

The online methods we found show the best quality of image vectorization. 
However, this is achieved by using a large number of shapes, the number of which can only be controlled indirectly by specifying the number of available colors. The results are presented in Tab.~\ref{tab:res}.

\begin{table}
\begin{center}
\begin{tabular}{| l | c | c | c | c | c|}
\hline
Method & Approach Type & Code & Versatility & Speed & Number of figures\\
\hline
DiffVG & ML: Iterative & + & + & Low & User-defined \\
Im2Vec & ML: VAE & + & - & Pretrain needed & User-defined \\
LIVE & ML: Iterative & + & + & Very low & User-defined \\
ClipGEN & ML: Iterative+DL & - & - & Pretrain needed & User-defined \\
Mang2Vec & ML: RL & + & - & Medium & Many \\
DVoTD & ML: DL & + & - & Medium & Many \\
VTracer & Algorithmic, online & + & + & Very High & Medium\\
\hline
\end{tabular}
\end{center}
\caption{Classification and comparison of vectorization methods. 'ML' means machine learning, 'DL' means deep learning, and 'RL' means reinforcement learning.}\label{tab:res}
\end{table}

However, none of the considered methods could recreate exact copies using such a number of figures.

\section{Conclusion}
In this work we have shown that current image vectorization methods are difficult to use in practice. 
Online methods without manual hyperparameter tuning create images containing a large number of paths, which increases the amount of used memory, and design refinement becomes a time-consuming task. 
All the existing machine learning-compatible methods also require human control and adjustment of method iterations number, output vector image parameters, etc. 
The Im2Vec model is not capable of storing and generating complex images and is not a universal vectorizer that could create a vector analog for any input image. 
The LIVE method is the only universal model that allows you to control the number of drawn shapes, however, due to the use of an iterative approach, generating a single image takes a huge amount of time.

According to our measurements, DiffVG is the fastest among ML methods without much quality losses. 
However, a large number of paths are required for high-quality results. 
At the same time, LIVE is able to get no worse quality  using fewer shapes. 
However, the main problem with the LIVE method is its extra-long running time.

Perhaps, generally, it is best to use online methods, but they do not allow you to adjust the number of applied shapes.

% The main outcome is that ...
To summarize, vectorization methods are connected with a tradeoff between image quality, path number, segment number, closed or not paths are, number of iterations, and running time.

\bibliographystyle{splncs04}
\bibliography{egbib}





% LNCS Homepage, \url{http://www.springer.com/lncs}. Last accessed 4
% \end{thebibliography}
\end{document}